\PassOptionsToPackage{numbers}{natbib}
\documentclass{winnower}

\usepackage[utf8]{inputenc} 
\usepackage[T1]{fontenc}    
\usepackage{hyperref}       
\usepackage{url}            
\usepackage{booktabs}       
\usepackage{amsfonts}       
\usepackage{nicefrac}       
\usepackage{microtype}      
\usepackage{amssymb}
\usepackage{amsmath,bm}
\usepackage{graphicx,epstopdf}
\usepackage{multirow}
\usepackage{caption}
\usepackage{subfigure}
\usepackage{comment}
\usepackage{float}
\usepackage{lscape}

\usepackage{color}
\usepackage{amssymb,dsfont}
\usepackage{ulem}
\usepackage{epstopdf}
\usepackage{leftidx}
\usepackage{multirow}
\usepackage{bigdelim}
\usepackage{amsfonts}
\usepackage{mathrsfs}
\usepackage{blkarray,adjustbox}
\usepackage{lineno}
\usepackage{enumitem}
\usepackage{mathpazo}
\usepackage{booktabs} 
\usepackage{float}
\usepackage{amsmath}
\usepackage{hyperref}
\usepackage[dvipsnames]{xcolor}

\newtheorem{Theorem}{Theorem}[section]

\newtheorem{Remark}{Remark}[section]
\newtheorem{Definition}{Definition}[section]

\newcommand{\etr}{\mathrm{etr}}
\newcommand{\tr}{\mathrm{tr}}

\graphicspath{{./Figures/}} 

\begin{document}

\title{Multivariate Gaussian and Student$-t$ process regression for
multi-output prediction}

\author[1]{Zexun Chen \footnote{Corresponding author: Zexun Chen, Email: sxtpy2010@gmail.com}}
\author[1]{Bo Wang}
\author[1]{Alexander N. Gorban}
\affil[1]{Department of Mathematics \\
University of Leicester, Leicester, LE1 7RH, UK}

\date{}

\maketitle

\begin{abstract}
Gaussian process model for vector-valued function has been shown to be useful
for multi-output prediction. The existing method for this model is to re-formulate the matrix-variate Gaussian distribution as a multivariate normal distribution. Although it is effective in many cases, re-formulation is not always workable
and is difficult to apply to other distributions because not all matrix-variate distributions can be transformed to respective multivariate distributions,
such as the case for matrix-variate Student$-t$ distribution.
In this paper, we propose a unified framework which
is used not only to
introduce a novel multivariate Student$-t$ process regression model
(MV-TPR) for multi-output prediction, but also to reformulate the multivariate Gaussian process regression (MV-GPR) that overcomes some limitations of the existing methods.
Both MV-GPR and MV-TPR have closed-form expressions for the marginal likelihoods and predictive distributions under this unified framework and thus
can adopt the same optimization approaches as used in the conventional GPR.
The usefulness of the proposed methods is illustrated through several simulated and real data examples. In particular, we verify empirically that MV-TPR has superiority for the datasets considered, including air quality prediction and bike rent prediction. At last, the proposed methods are shown to produce profitable investment strategies in the stock markets.
\end{abstract}

\section{Introduction}\label{intro}
Over the last few decades, Gaussian processes regression (GPR) has been proven to be a powerful and effective method for non-linear regression problems due to many favorable properties, such as simple structure of obtaining and expressing uncertainty in predictions, the capability of capturing a wide variety of behaviour by parameters, and a natural Bayesian interpretation \cite{cite:RCE,boyle2005dependent}. In 1996,
Neal \cite{cite:neal} revealed that many Bayesian regression models based on neural network converge to Gaussian processes (GP) in the limit of an infinite number of hidden units \cite{williams1997computing}. GP has been suggested as a replacement for supervised neural networks in non-linear regression \cite{cite:MacKay,cite:WCKI} and classification \cite{cite:MacKay}. Furthermore, GP has excellent capability for forecasting time series \cite{cite:BB,cite:BBS}.

Despite the popularity of GPR in various modelling tasks, there still exists a conspicuous imperfection, that is, the majority of GPR models are implemented for single response variables or considered independently for multiple responses variables without consideration of their correlation \cite{boyle2005dependent,wang2015gaussian}. In order to resolve the multi-output prediction problem, Gaussian process regression for vector-valued function is proposed and regarded as a pragmatic and straightforward method. The core of this method is to vectorize the multi-response variables and construct a "big" covariance, which describes the correlations between the inputs as well as
between the outputs \cite{boyle2005dependent,wang2015gaussian,chakrabarty2015bayesian,alvarez2011kernels}. This modelling strategy is feasible due to the fact that the matrix-variate Gaussian distributions can be re-formulated as multivariate Gaussian distributions \cite{chakrabarty2015bayesian,gupta1999matrix}. Intrinsically, Gaussian process regression for vector-valued function is still a conventional Gaussian process regression model since
it merely vectorizes multi-response variables, which are assumed to follow a developed case of GP with a reproduced kernel.
As an extension, it is natural to consider more general elliptical processes models for multi-output prediction. However, the vectorization method cannot be used to extend multi-output GPR because the equivalence between vectorized matrix-variate and multivariate distributions only exists in Gaussian cases \cite{gupta1999matrix}.

To overcome this drawback, in this paper we propose a unified framework which:
(1) is used to
introduce a novel multivariate Student$-t$ process regression model
(MV-TPR) for multi-output prediction, (2) is used to reformulate the multivariate Gaussian process regression (MV-GPR) that overcomes some limitations of the existing methods, (3) can be used to derive regression models of
general elliptical processes.
Both MV-GPR and MV-TPR have closed-form expressions for the marginal likelihoods and predictive distributions under this unified framework and thus
can adopt the same optimization approaches as used in the conventional GPR.
The usefulness of the proposed methods is illustrated through several simulated examples. Furthermore, we also verify empirically that MV-TPR has superiority in the prediction based on some widely-used datasets, including air quality prediction and bike rent prediction. The proposed methods are then applied to stock market modelling which shows that the profitable stock investment strategies can be
obtained.


The rest of the paper is organised as follows. Section 2 introduces some preliminaries of matrix-variate Gaussian and Student$-t$ distributions with their useful properties.
Section 3 presents the unified framework to reformulate
the multivariate Gaussian process regression and to derive the new
multivariate Student$-t$ process regression models. 
Some numerical experiments by the simulated data and real data
and the applications to stock market investment are presented in Section 4.
Conclusion and discussion are given in Section 5.

\section{Backgrounds and notations}\label{sec:preliminaries}
Matrix-variate Gaussian and Student$-t$ distributions have many useful properties, as discussed in the literatures \cite{gupta1999matrix,dawid1981some,zhu2008predictive}. For completeness
and easy referencing, below we list some of them which will be used in this paper.
\subsection{Matrix-variate Gaussian distribution}
\begin{Definition}\label{def:MV-G}
  The random matrix $X \in \mathds{R}^{n \times d}$ is said to have a matrix-variate Gaussian distribution with mean matrix $M \in \mathds{R}^{n \times d} $ and covariance matrix $\Sigma \in \mathds{R}^{n \times n},\Omega \in \mathds{R}^{d \times d}$ if and only if its probability density function is given by
  \begin{equation}\label{eq:matrixGaussian}
    p(X|M,\Sigma,\Omega) = (2\pi)^{-\frac{dn}{2}}\det(\Sigma)^{-\frac{d}{2}}\det(\Omega)^{-\frac{n}{2}}\etr(-\frac{1}{2}\Omega^{-1}(X-M)^{\mathrm{T}}\Sigma^{-1}(X -M)),
  \end{equation}
where $\etr(\cdot)$ is exponential of matrix trace, $\Omega$ and $\Sigma$ are positive semi-definite. It is denoted $X \sim \mathcal{MN}_{n,d}(M, \Sigma, \Omega)$. Without loss of clarity, it is denoted $X \sim \mathcal{MN}(M, \Sigma, \Omega)$.
\end{Definition}
Like multivariate Gaussian distribution, matrix-variate Gaussian distribution also possesses several important properties as follows.
\begin{Theorem}[Transposable]\label{thm:transponsable-G}
  If $X \sim \mathcal{MN}_{n,d}(M, \Sigma, \Omega)$, then $X^{\mathrm{T}} \sim \mathcal{MN}_{d,n}(M^{\mathrm{T}}, \Omega, \Sigma)$.
\end{Theorem}
The matrix-variate Gaussian is related to the multivariate Gaussian in the following way.
\begin{Theorem}[Vectorizable]\label{thm:vectorizable-G}
  $X \sim \mathcal{MN}_{n,d}(M, \Sigma, \Omega)$ if and only if
  $$\mathrm{vec}(X^{\mathrm{T}}) \sim \mathcal{N}_{nd}(\mathrm{vec}(M^{\mathrm{T}}), \Sigma \otimes \Omega ),$$
  where $\mathrm{vec}(\cdot)$ is vector operator and $\otimes$ is Kronecker product (or called tensor product).
\end{Theorem}
Furthermore, the matrix-variate Gaussian distribution is consistent under the marginalization and conditional distribution.
\begin{Theorem}[Marginalization and conditional distribution]\label{thm:MarginCondition-G}
  Let $X \sim \mathcal{MN}_{n,d}(M, \Sigma, \Omega)$, and partition $X, M, \Sigma$ and $\Omega$ as
\begin{equation*}
  X =
  \begin{adjustbox}{raise=-1ex} $\displaystyle
    \begin{blockarray}{[c]c}
      X_{1r} & n_1\\
      X_{2r} & n_2
    \end{blockarray} $
  \end{adjustbox}
  =
    \begin{adjustbox}{raise=-1ex} $\displaystyle
    \begin{blockarray}{cc}
          &            \\
      \begin{block}{[cc]}
         X_{1c} & X_{2c} \\
      \end{block}
      d_1 & d_2
    \end{blockarray}$
  \end{adjustbox}
  ,\quad
  M =
  \begin{adjustbox}{raise=-1ex} $\displaystyle
    \begin{blockarray}{[c]c}
      M_{1r} & n_1\\
      M_{2r} & n_2
    \end{blockarray} $
  \end{adjustbox}
  =
    \begin{adjustbox}{raise=-1ex} $\displaystyle
    \begin{blockarray}{cc}
          &        \\
      \begin{block}{[cc]}
         M_{1c} & M_{2c} \\
      \end{block}
      d_1 & d_2
    \end{blockarray} $
  \end{adjustbox}
\end{equation*}
\begin{equation*}
  \Sigma =
  \begin{adjustbox}{raise=-2.5ex} $\displaystyle
    \begin{blockarray}{ccc}
      \begin{block}{[cc]c}
        \Sigma_{11} & \Sigma_{12} &n_1\\
        \Sigma_{21} & \Sigma_{22} & n_2\\
      \end{block}
      n_1 & n_2 &
    \end{blockarray} $
  \end{adjustbox}
  \quad \text{and} \quad
  \Omega =
  \begin{adjustbox}{raise=-2.5ex} $\displaystyle
    \begin{blockarray}{ccc}
      \begin{block}{[cc]c}
        \Omega_{11} & \Omega_{12} & d_1\\
        \Omega_{21} & \Omega_{22} & d_2\\
      \end{block}
      d_1 & d_2 &
    \end{blockarray} $
  \end{adjustbox},
\end{equation*}
where $n_1,n_2, d_1,d_2$ is the column or row length of the corresponding vector or matrix. Then,
\begin{enumerate}[leftmargin=*,labelsep=3mm]
   \item $X_{1r} \sim \mathcal{MN}_{n_1,d}\left(M_{1r},\Sigma_{11},\Omega \right)$,
   $$
   X_{2r}|X_{1r} \sim \mathcal{MN}_{n_2,d}\left(M_{2r} + \Sigma_{21}\Sigma_{11}^{-1}(X_{1r}-M_{1r}),\Sigma_{22\cdot1},\Omega \right);
   $$
   \item $X_{1c} \sim \mathcal{MN}_{n,d_1}\left(M_{1c},\Sigma,\Omega_{11}\right)$,
   $$
       X_{2c}|X_{1c} \sim \mathcal{MN}_{n,d_2}\left(M_{2c} + (X_{1c}-M_{1c})\Omega_{11}^{-1}\Omega_{12},\Sigma,\Omega_{22\cdot1} \right);
   $$
\end{enumerate}
where $\Sigma_{22\cdot1} $and $\Omega_{22\cdot1}$ are the Schur complement \cite{zhang2006schur} of $\Sigma_{11}$ and $\Omega_{11}$, respectively,
$$
\Sigma_{22\cdot1} = \Sigma_{22} - \Sigma_{21}\Sigma_{11}^{-1}\Sigma_{12} , \quad \Omega_{22\cdot1} = \Omega_{22} - \Omega_{21}\Omega_{11}^{-1}\Omega_{12}.
$$
\end{Theorem}
\subsection{Matrix-variate Student\texorpdfstring{$-t$}{t} distribution}
\begin{Definition}\label{def:MV-T}
  The random matrix $X \in \mathds{R}^{n \times d}$ is said to have a matrix-variate Student$-t$ distribution with the mean matrix $M\in \mathds{R}^{n \times d}$ and covariance matrix $\Sigma \in \mathds{R}^{n \times n},\Omega \in \mathds{R}^{d \times d}$ and the degree of freedom $\nu$ if and only if the probability density function is given by
\begin{eqnarray}\label{eq:MatrixT}
p(X|\nu, M, \Sigma, \Omega)&=&\frac{\Gamma_n [\frac{1}{2}(\nu + d + n -1)]}{\pi^{\frac{1}{2}dn}\Gamma_n [\frac{1}{2}(\nu + n -1)]}\det(\Sigma)^{-\frac{d}{2}} \det(\Omega)^{-\frac{n}{2}} \times  \nonumber \\
  && \qquad \quad \det (\mathbf{I}_n + \Sigma^{-1}(X-M)\Omega^{-1}(X-M)^{\mathrm{T}})^{-\frac{1}{2}(\nu + d + n -1)},
\end{eqnarray}
where $\Omega$ and $\Sigma$ are positive semi-definite, and
$$
\Gamma_n(\lambda) = \pi^{n(n-1)/4}\prod_{i=1}^{n}\Gamma(\lambda + \frac{1}{2} - \frac{i}{2}).
$$
We denote this by $X \sim \mathcal{MT}_{n,d}(\nu, M, \Sigma, \Omega)$. Without loss of clarity, it is denoted $X \sim \mathcal{MT}(\nu, M, \Sigma, \Omega)$.
\end{Definition}
\begin{Theorem}[Expectation and covariance]
  Let $X \sim \mathcal{MT}(\nu, M, \Sigma, \Omega)$, then
  $$ \mathbb{E}(X) = M,\quad \mathrm{cov}(\mathrm{vec}(X^{\mathrm{T}})) = \frac{1}{\nu -2}\Sigma \otimes \Omega , \nu > 2.$$
\end{Theorem}
\begin{Theorem}[Transposable]
  If $X \sim \mathcal{MT}_{n,d}(\nu,M, \Sigma, \Omega)$, then
  $X^{\mathrm{T}} \sim \mathcal{MT}_{d,n}(\nu,M^{\mathrm{T}}, \Omega,\Sigma).$
\end{Theorem}
\begin{Theorem}[Asymptotics]\label{thm:Asymptotics-T}
  Let $X \sim \mathcal{MT}_{n,d}(\nu,M, \Sigma, \Omega)$,then $X \overset{d}{\to} \mathcal{MN}_{n,d}(M, \Sigma, \Omega)$ as $\nu \to \infty$, where "$\overset{d}{\to}$" denotes convergence in distribution.
\end{Theorem}
\begin{Theorem}[Marginalization and conditional distribution]\label{thm:MarginCondition-T}
  Let $X \sim \mathcal{MT}_{n,d}(\nu,M, \Sigma, \Omega)$, and partition $X, M, \Sigma$ and $\Omega$ as
\begin{equation*}
  X =
  \begin{adjustbox}{raise=-1ex} $\displaystyle
    \begin{blockarray}{[c]c}
      X_{1r} & n_1\\
      X_{2r} & n_2
    \end{blockarray} $
  \end{adjustbox}
  =
    \begin{adjustbox}{raise=-1ex} $\displaystyle
    \begin{blockarray}{cc}
    & \\
      \begin{block}{[cc]}
         X_{1c} & X_{2c} \\
      \end{block}
      d_1 & d_2
    \end{blockarray}$
  \end{adjustbox}
  ,\quad
  M =
  \begin{adjustbox}{raise=-1ex} $\displaystyle
    \begin{blockarray}{[c]c}
      M_{1r} & n_1\\
      M_{2r} & n_2
    \end{blockarray} $
  \end{adjustbox}
  =
    \begin{adjustbox}{raise=-1ex} $\displaystyle
    \begin{blockarray}{cc}
    & \\
      \begin{block}{[cc]}
         M_{1c} & M_{2c} \\
      \end{block}
      d_1 & d_2
    \end{blockarray} $
  \end{adjustbox}
\end{equation*}
\begin{equation*}
  \Sigma =
  \begin{adjustbox}{raise=-2.5ex} $\displaystyle
    \begin{blockarray}{ccc}
      \begin{block}{[cc]c}
        \Sigma_{11} & \Sigma_{12} &n_1\\
        \Sigma_{21} & \Sigma_{22} & n_2\\
      \end{block}
      n_1 & n_2 &
    \end{blockarray} $
  \end{adjustbox}
  \quad \text{and} \quad
  \Omega =
  \begin{adjustbox}{raise=-2.5ex} $\displaystyle
    \begin{blockarray}{ccc}
      \begin{block}{[cc]c}
        \Omega_{11} & \Omega_{12} & d_1\\
        \Omega_{21} & \Omega_{22} & d_2\\
      \end{block}
      d_1 & d_2 &
    \end{blockarray} $
  \end{adjustbox},
\end{equation*}
where $n_1,n_2, d_1,d_2$ is the column or row length of the corresponding vector or matrix. Then,
\begin{enumerate}[leftmargin=*,labelsep=3mm]
   \item $X_{1r} \sim \mathcal{MT}_{n_1,d}\left(\nu, M_{1r},\Sigma_{11},\Omega \right)$,
   \begin{align*}
     X_{2r}|X_{1r} \sim &  \mathcal{MT}_{n_2,d} \Big( \nu + n_1, M_{2r} + \Sigma_{21}\Sigma_{11}^{-1}(X_{1r}-M_{1r}),\Sigma_{22\cdot1}, \\
       & \Omega  + (X_{1r}-M_{1r})^{\mathrm{T}}\Sigma_{11}^{-1}(X_{1r}-M_{1r}) \Big);
   \end{align*}
   \item $X_{1c} \sim \mathcal{MT}_{n,d_1}\left(,\nu, M_{1c},\Sigma,\Omega_{11}\right)$,
   \begin{align*}
     X_{2c}|X_{1c} \sim  & \mathcal{MT}_{n,d_2} \Big(\nu + d_1, M_{2c} + (X_{1c}-M_{1c})\Omega_{11}^{-1}\Omega_{12}, \\
       & \Sigma + (X_{1c}-M_{1c})\Omega_{11}^{-1}(X_{1c}-M_{1c})^{\mathrm{T}},\Omega_{22\cdot1} \Big);
   \end{align*}
\end{enumerate}
where $\Sigma_{22\cdot1} $and $\Omega_{22\cdot1}$ are the Schur complement of $\Sigma_{11}$ and $\Omega_{11}$, respectively,
$$
\Sigma_{22\cdot1} = \Sigma_{22} - \Sigma_{21}\Sigma_{11}^{-1}\Sigma_{12} , \quad \Omega_{22\cdot1} = \Omega_{22} - \Omega_{21}\Omega_{11}^{-1}\Omega_{12}.
$$
\end{Theorem}

\begin{Remark}\label{rmk:remark:section2}
It can be seen that matrix-variate Student$-t$ distribution has many properties
similar to matrix-variate Gaussian distribution, and it converges to
matrix-variate Gaussian distribution if its degree of freedom tends to infinity. However, matrix-variate Student$-t$ distribution lacks the property of
vectorizability (Theorem \ref{thm:vectorizable-G}) \cite{gupta1999matrix}.
As a consequence Student$-t$ process regression for multiple outputs
cannot be derived by vectorizing the multi-response variables.
In the next section, we propose a new framework to introduce multivariate
Student$-t$ process regression model.
\end{Remark}

\section{Multivariate Gaussian and Student$-t$ process regression models}\label{models}

\subsection{Multivariate Gaussian process regression (MV-GPR)}

If $\bm{f}$ is a multivariate Gaussian process on $\mathcal{X}$ with vector-valued mean function $\bm{u} : \mathcal{X}\mapsto \mathds{R}^d$, covariance function (also called kernel) $k: \mathcal{X}\times \mathcal{X} \mapsto \mathds{R}$ and positive semi-definite parameter matrix $\Omega \in \mathds{R}^{d \times d}$, then any finite collection of vector-valued variables have a joint matrix-variate Gaussian distribution,
 $$
 [\bm{f}(x_1)^{\mathrm{T}},\ldots,\bm{f}(x_n)^{\mathrm{T}}]^{\mathrm{T}} \sim \mathcal{MN}(M, \Sigma, \Omega),n \in\mathds{N},
 $$
 where $\bm{f}, \bm{u} \in \mathds{R}^d$ are row vectors whose components are the functions $\{f_i\}_{i=1}^d$ and $\{\mu_i\}_{i=1}^d$ respectively. Furthermore, $M \in \mathds{R}^{n \times d}$ with $M_{ij} = \mu_{j}(x_i)$, and $\Sigma \in \mathds{R}^{n \times n}$ with $\Sigma_{ij} = k(x_i,x_j)$. Sometimes $\Sigma$ is called column covariance matrix while $\Omega$ is row covariance matrix. We denote $\bm{f} \sim \mathcal{MGP}(\bm{u}, k, \Omega)$.


In conventional GPR methods, the noisy model $y=f(x)+\varepsilon$ is usually considered. However, for Student$-t$ process regression such a model
is analytically intractable \cite{shah2014student}. Therefore
we adopt the method used in \cite{shah2014student}
and consider the noise-free regression model where the noise term
is incorporated into the kernel function.

Given $n$ pairs of observations $\{(x_i,\bm{y}_i)\}_{i=1}^n, x_i \in \mathds{R}^p, \bm{y}_i \in \mathds{R}^{1\times d}$, we assume the following model
\begin{eqnarray*}
  \bm{f} & \sim & \mathcal{MGP}(\bm{u},k',\Omega), \\
   \bm{y}_i & = & \bm{f}(x_i), \mbox{ for} \ i = 1,\cdots,n,
\end{eqnarray*}
where
\begin{equation}
k' = k(x_i,x_j) + \delta_{ij}\sigma_n^2, \label{kenerl-with-noise}
\end{equation}
and $\delta_{ij}=1$ if $i=j$, otherwise $\delta_{ij}=0$.
Note that the second term in \eqref{kenerl-with-noise} represents
the random noises.

We assume $\bm{u} = \bm{0}$ as commonly done in GPR.
By the definition of multivariate Gaussian process, it yields that the collection of functions $[\bm{f}(x_1),\ldots,\bm{f}(x_n)]$ follow a matrix-variate Gaussian distribution
$$
[\bm{f}(x_1)^{\mathrm{T}},\ldots,\bm{f}(x_n)^{\mathrm{T}}]^{\mathrm{T}} \sim \mathcal{MN}(\bm{0},K',\Omega),
$$
where $K'$ is the $n \times n$ covariance matrix of which the $(i,j)$-th element $[K']_{ij} = k'(x_i,x_j)$.

To predict a new variable $\bm{f}_* = [f_{*1},\ldots,f_{*m}]^\mathrm{T}$ at the test locations $X_* = [x_{n+1},\ldots,x_{n+m}]^\mathrm{T}$, the joint distribution of the training observations $Y = [\bm{y}_1^{\mathrm{T}},\cdots,\bm{y}_n^{\mathrm{T}}]^{\mathrm{T}}$ and the predictive targets $\bm{f}_*$ are given by
\begin{equation}\label{joint}
  \begin{bmatrix}
  Y  \\
  \bm{f}_*
  \end{bmatrix} \sim
  \mathcal{MN}    \left(
  \bm{0},
  \begin{bmatrix}
  K'(X,X)   \quad K'(X_*,X)^{\mathrm{T}} \\
  K'(X_*,X) \ \ K'(X_*,X_*)
  \end{bmatrix},
  \Omega   \right),
\end{equation}
where $K'(X,X)$ is an $n \times n$ matrix of which the $(i,j)$-th element
$[K'(X,X)]_{ij} = k'(x_{i},x_j)$, $K'(X_*,X)$ is an $m \times n$ matrix of which the $(i,j)$-th element
$[K'(X_*,X)]_{ij} = k'(x_{n+i},x_j)$, and $K'(X_*,X_*)$ is an $m \times m$ matrix with
 the $(i,j)$-th element $[K'(X_*,X_*)]_{ij} = k'(x_{n+i},x_{n+j})$. Thus, taking advantage of conditional distribution of multivariate Gaussian process, the predictive distribution is
 \begin{equation}
   p(\bm{f}_*|X,Y,X_*) =    \mathcal{MN}(\hat{M},\hat{\Sigma},\hat{\Omega}),
 \end{equation}
 where
\begin{eqnarray}
  \hat{M}  &= &K'(X_*,X)^{\mathrm{T}}K'(X,X)^{-1}Y, \\
  \hat{\Sigma}  &=& K'(X_*,X_*)  - K'(X_*,X)^{\mathrm{T}}K'(X,X)^{-1}K'(X_*,X),\\
  \hat{\Omega} &= & \Omega  .
\end{eqnarray}
Additionally, the expectation and the covariance are obtained,
\begin{eqnarray}
  \mathbb{E}[\bm{f}_*] &=& \hat{M}=K'(X_*,X)^{\mathrm{T}}K'(X,X)^{-1}Y, \label{pred_mean}\\
  \mathrm{cov}(\mathrm{vec}(\bm{f}^{\mathrm{T}}_*)) &=& \hat{\Sigma}\otimes \hat{\Omega}  = [K'(X_*,X_*)  - K'(X_*,X)^{\mathrm{T}}K'(X,X)^{-1}K'(X_*,X)]\otimes \Omega . \label{pred_var}
\end{eqnarray}

%
%

\subsubsection{Kernel}

Although there are two covariance matrices in the above regression model,
the column covariance
and the row covariance, only the column covariance depends on inputs
and is considered as kernel since it
contains our presumptions about the function we wish to learn and define the closeness and similarity between data points \cite{cite:Rbook}.
As in conventional GPR, the choice of kernels has a profound impact on the performance of multivariate Gaussian process regression
(as well as multivariate Student$-t$ process regression introduced later).

A wide range of useful kernels have been proposed in the literature,
such as linear, rational quadratic, and Mat{\'e}rn \cite{cite:Rbook}.
But the Squared Exponential (SE) kernel is the most commonly used
due to its simple form and many desirable properties such as
smoothness and integrability with other functions, although
it could oversmooth the data, especially financial data. 

The Squared Exponential (SE) kernel is defined as
$$
k_{SE}(x,x') = s_f^2 \exp (- \frac{\|x-x'\|^2}{2\ell^2}),
$$
where $s_f^2$ is the signal variance and can also be considered as an output-scale amplitude and the parameter $\ell$ is the input (length or time) scale \cite{cite:twins}.
The kernel can also be defined by Automatic Relevance Determination (ARD)
$$
k_{SEard}(\bm{x},\bm{x}') = s_f^2 \exp (- \frac{(\bm{x}-\bm{x}')^{\mathrm{T}}\Theta^{-1}(\bm{x}-\bm{x}')}{2}),
$$
where $\Theta$ is a diagonal matrix with the element components $\{\ell_i^2\}_{i=1}^p$, which represents the length scales for each corresponding input dimension.

For convenience and the purpose of demonstration, SE kernel is used in all our experiments where there is only one input variable whilst
SEard is used in those with multiple input variables.
It should be noted that there is no technical difficulty to
use other kernels in our models.



\subsubsection{Parameter estimation}
The hyper-parameters involved in the kernel of MV-GPR need to be estimated from the training data. Although Monte Carlo methods can perform GPR without the need of estimating hyperparameters \cite{cite:WCKI,cite:BBS,cite:monte,mackay1998introduction}, the common approach is
to estimate them by means of maximum marginal likelihood due to the high computational cost of Monte Carlo methods. $\Omega$ is an extra parameter compared to the conventional GPR model, hence
the unknown parameters include the hyper-parameters in the kernel, the noise variance $\sigma_n^2$ and the row covariance parameter matrix $\Omega$.

Because $\Omega$ is positive semi-definite, it can be denoted as $\Omega = \Phi \Phi^{\mathrm{T}}$, where
$$
\Phi =
\left[
\begin{matrix}
 \phi_{11}      & 0      & \cdots & 0      \\
 \phi_{21}      & \phi_{22}      & \cdots & 0      \\
 \vdots & \vdots & \ddots & \vdots \\
 \phi_{d1}      & \phi_{d2}      & \cdots & \phi_{dd}      \\
\end{matrix}
\right].
$$
To guarantee the uniqueness of $\Phi$, the diagonal elements are restricted to be positive and denote $\varphi_{ii} = \ln(\phi_{ii})$ for $i = 1,2,\cdots,d$.

In MV-GPR model, the observations follow a matrix-variate Gaussian distribution $Y \sim \mathcal{MN}_{n,d}(\bm{0},K',\Omega)$ where
$K'$ is the noisy column covariance matrix with element $[K']_{ij} = k'(x_i,x_j)$ so that $K' = K + \sigma_n^2 \mathbf{I}$ where $K$ is noise-free column covariance matrix with element $[K]_{ij} = k(x_i,x_j)$. As we know there are hyper-parameters in the kernel $k$ so that we can denote $K = K_{\theta}$. The hyper-parameter set denotes $\Theta = \{\theta_1, \theta_2, \ldots\}$, thus
$$
\frac{\partial K'}{\partial \sigma_n^2} = \mathbf{I}_n,  \quad \frac{\partial K'
}{\partial \theta_i} = \frac{\partial K_{\theta}}{\partial \theta_i}
$$

According to the matrix-variate distribution,  the negative log marginal likelihood of observations is
\begin{equation}\label{matrixLikelihood}
  \mathcal{L} = \frac{nd}{2}\ln(2\pi) + \frac{d}{2}\ln \det(K') + \frac{n}{2}\ln \det(\Omega) + \frac{1}{2}\tr((K')^{-1}Y\Omega^{-1}Y^{\mathrm{T}}).
\end{equation}

The derivatives of the negative log marginal likelihood with respect to parameter $\sigma_n^2$, $\theta_i$, $\phi_{ij}$ and $\varphi_{ii}$ are as follows
\begin{align*}
  \frac{\partial \mathcal{L}}{\partial \sigma_n^2} &=  \frac{d}{2}\tr((K')^{-1}) - \frac{1}{2}\tr(\alpha_{K'}\Omega^{-1}\alpha_{K'}^{\mathrm{T}}),\\
  \frac{\partial \mathcal{L}}{\partial \theta_i} & = \frac{d}{2}\tr\left((K')^{-1}\frac{\partial K_{\theta}}{\partial \theta_i}\right) - \frac{1}{2}\tr\left(\alpha_{K'}\Omega^{-1}\alpha_{K'}^{\mathrm{T}}\frac{\partial K_{\theta}}{\partial \theta_i}\right), \\
   \frac{\partial \mathcal{L}}{\partial \phi_{ij}} & = \frac{n}{2}\tr[\Omega^{-1}(\mathbf{E}_{ij}\Phi^{\mathrm{T}} + \Phi \mathbf{E}_{ij})] - \frac{1}{2}\tr[\alpha_{\Omega}(K')^{-1}\alpha_{\Omega}^{\mathrm{T}}(\mathbf{E}_{ij}\Phi^{\mathrm{T}} + \Phi \mathbf{E}_{ij})], \\
  \frac{\partial \mathcal{L}}{\partial \varphi_{ii}} & = \frac{n}{2}\tr[\Omega^{-1}(\mathbf{J}_{ii}\Phi^{\mathrm{T}} + \Phi \mathbf{J}_{ii})] - \frac{1}{2}\tr[\alpha_{\Omega}(K')^{-1}\alpha_{\Omega}^{\mathrm{T}}(\mathbf{J}_{ii}\Phi^{\mathrm{T}} + \Phi \mathbf{J}_{ii})],
\end{align*}
where $\alpha_{K'} = (K')^{-1}Y$, $\alpha_{\Omega} = \Omega^{-1}Y^{\mathrm{T}}$, $\mathbf{E}_{ij}$ is the $d \times d$ elementary matrix having unity in the (i,j)-th element and zeros elsewhere, and $\mathbf{J}_{ii}$ is the same as $\mathbf{E}_{ij}$ but with the unity being replaced by $e^{\varphi_{ii}}$.
The details are provided in \ref{app:MV-GPR-Gradient}.

Hence standard gradient-based numerical optimisation techniques, such as Conjugate Gradient method, can be used to minimise the negative log marginal
likelihood function to obtain the estimates of the parameters.
Note that since the random noise is incorporated into the kernel function the noise variance is estimated alongside the other hyper-parameters.

\subsubsection{Comparison with the existing methods}

Compared with the existing multi-output GPR methods \cite{alvarez2011kernels,boyle2005dependent,wang2015gaussian},
our proposed method possesses several advantages.

Firstly, the existing methods have to vectorize the multi-output matrix
in order to utilise the GPR models. It is complicated and not always workable
if the numbers of outputs and observations are large.
In contrast, our proposed MV-GPR has more straightforward form where the model settings, derivations and computations are all directly performed in matrix form.
In particular, we use column covariance (kernel) and row covariance
to capture all the correlations together in the multivariate outputs,
rather than assuming a separate kernel for each output and
constructing a “big” covariance matrix by Kronecker product as done
in \cite{boyle2005dependent}.


Secondly, the existing methods rely on the equivalence between vectorized
matrix-variate Gaussian distribution and multivariate Gaussian distribution.
However, this equivalence does not exist for other elliptical distributions
such as matrix-variate Student$-t$ distribution \cite{gupta1999matrix}.
Therefore, the existing methods for multi-output Gaussian process regression
 cannot be applied to Student$-t$ process regression.
On the other hand, our proposed MV-GPR is based on matrix forms directly
 and does not require vectorization so it can naturally be extended to MV-TPR
 as we will do in the next subsection.

Therefore, our proposed MV-GPR provides not only a new derivation of the multi-output Gaussian process regression model, but also a unified framework to
derive more general elliptical processes models.

\subsection{Multivariate Student\texorpdfstring{$-t$}{t} process regression
(MV-TPR)}

In this subsection we propose a new nonlinear regression model for multivariate response,
namely multivariate Student\texorpdfstring{$-t$}{t} process regression model (MV-TPR),
using the framework discussed in the previous subsections.
MV-TPR is an extension to multi-output GPR, as well as an extension to the
univariate Student$-t$ process
regression proposed in \cite{shah2014student}.

By definition, if
$\bm{f}$ is a multivariate Student$-t$ process on $\mathcal{X}$ with parameter $\nu>2$, vector-valued mean function $\bm{u} : \mathcal{X}\mapsto \mathds{R}^d$, covariance function (also called kernel) $k: \mathcal{X}\times \mathcal{X} \mapsto \mathds{R}$ and positive semi-definite parameter matrix $\Omega \in \mathds{R}^{d \times d}$, then any finite collection of vector-valued variables have a joint matrix-variate Student$-t$ distribution,
 $$
 [\bm{f}(x_1)^{\mathrm{T}},\ldots,\bm{f}(x_n)^{\mathrm{T}}]^{\mathrm{T}} \sim \mathcal{MT}(\nu, M, \Sigma, \Omega),n \in\mathds{N},
 $$
 where $\bm{f}, \bm{u} \in \mathds{R}^d$ are row vectors whose components are the functions $\{f_i\}_{i=1}^d$ and $\{\mu_i\}_{i=1}^d$ respectively. Furthermore, $M \in \mathds{R}^{n \times d}$ with $M_{ij} = \mu_{j}(x_i)$, and $\Sigma \in \mathds{R}^{n \times n}$ with $\Sigma_{ij} = k(x_i,x_j)$. We denote $\bm{f} \sim \mathcal{MTP}(\nu, \bm{u}, k, \Omega)$.


Therefore MV-TPR model can be formulated along the same line as MV-GPR based on the definition of multivariate Student$-t$ process. We present the model briefly below.

Given $n$ pairs of observations $\{(x_i,\bm{y}_i)\}_{i=1}^n, x_i \in \mathds{R}^p, \bm{y}_i \in \mathds{R}^{1\times d}$, we assume
\begin{eqnarray*}
  \bm{f} & \sim & \mathcal{MTP}(\nu,\bm{u},k',\Omega),\nu>2, \\
   \bm{y}_i & = & \bm{f}(x_i), \mbox{ for} \ i = 1,\cdots,n,
\end{eqnarray*}
where $\nu$ is the degree of freedom of Student$-t$ process and the remaining parameters have the same meaning of MV-GPR model. Consequently, the predictive distribution is obtained as
 \begin{equation}
   p(\bm{f}_*|X,Y,X_*) =    \mathcal{MT}(\hat{\nu},\hat{M},\hat{\Sigma},\hat{\Omega}),
 \end{equation}
 where
\begin{eqnarray}
  \hat{\nu} &=&\nu + n,\\
  \hat{M}&=&K'(X_*,X)^{\mathrm{T}}K'(X,X)^{-1}Y, \\
  \hat{\Sigma}  &=&K'(X_*,X_*)  - K'(X_*,X)^{\mathrm{T}}K'(X,X)^{-1}K'(X_*,X),\\
  \hat{\Omega} &=&\Omega + Y^{\mathrm{T}}K'(X,X)^{-1}Y.
\end{eqnarray}
According to the expectation and the covariance of matrix-variate Student$-t$ distribution, the predictive mean and covariance are given by
\begin{align}
  \mathbb{E}[\bm{f}_*] &= \hat{M}=K'(X_*,X)^{\mathrm{T}}K'(X,X)^{-1}Y, \\
  \mathrm{cov}(\mathrm{vec}(\bm{f}^{\mathrm{T}}_*)) &= \frac{1}{\nu +n-2}\hat{\Sigma}\otimes \hat{\Omega} \nonumber \\
   &= \frac{1}{\nu + n-2}[K'(X_*,X_*)  - K'(X_*,X)^{\mathrm{T}}K'(X,X)^{-1}K'(X_*,X)] \nonumber \\
   & \quad \otimes(\Omega + Y^{\mathrm{T}}K'(X,X)^{-1}Y) .
\end{align}


In the MV-TPR model, the observations are followed by a matrix-variate Student$-t$ distribution $Y \sim \mathcal{MT}_{n,d}(\nu,\bm{0},K',\Omega)$. The negative log marginal likelihood is
\begin{eqnarray*}\label{MultiLikelihoodT}
  \mathcal{L}  &=& \frac{1}{2}(\nu+ d+n -1) \ln \det(\mathbf{I}_n + (K')^{-1}Y\Omega^{-1}Y^{\mathrm{T}}) + \frac{d}{2}\ln \det(K') + \frac{n}{2}\ln \det(\Omega) \\
               & &  + \ln\Gamma_n \left(\frac{1}{2}(\nu + n -1)\right) - \ln \Gamma_n \left(\frac{1}{2}(\nu + d + n -1)\right) + \frac{1}{2}dn\ln\pi  \\
               &=& \frac{1}{2}(\nu+ d+n -1) \ln \det(K' +Y\Omega^{-1}Y^{\mathrm{T}}) - \frac{\nu + n -1}{2}\ln \det(K')  \\
               & & + \ln\Gamma_n \left(\frac{1}{2}(\nu + n -1)\right) - \ln \Gamma_n \left(\frac{1}{2}(\nu + d + n -1)\right)+ \frac{n}{2}\ln \det(\Omega)+ \frac{1}{2}dn\ln\pi .
\end{eqnarray*}

Therefore the parameters of MV-TPR contains all the parameters in MV-GPR and one more parameter: the degree of freedom $\nu$. The derivatives of the negative log marginal likelihood with respect to parameter $\nu$,$\sigma_n^2$, $\theta_i$, $\phi_{ij}$ and $\varphi_{ii}$ are as follows
\begin{align*}
  \frac{\partial \mathcal{L}}{\partial \nu} &= \frac{1}{2}\ln \det(U) - \frac{1}{2}\ln \det(K')  + \frac{1}{2}\psi_n(\frac{1}{2}\tau) - \frac{1}{2}\psi_n\left(\frac{1}{2}(\tau + d)\right), \\
  \frac{\partial \mathcal{L}}{\partial \sigma^2_n}
   & = \frac{(\tau+d)}{2}\tr(U^{-1}) - \frac{\tau}{2}\tr((K')^{-1}), \\
  \frac{\partial \mathcal{L}}{\partial \theta_i} & = \frac{(\tau+d)}{2}\tr\left(U^{-1} \frac{\partial K_{\theta}}{\partial \theta_i}\right) - \frac{\tau}{2}\tr\left(\Sigma^{-1} \frac{\partial K_{\theta}}{\partial \theta_i}\right),\\
    \frac{\partial \mathcal{L}}{\partial \phi_{ij}} & = - \frac{(\tau +d)}{2}\tr[U^{-1}\alpha_{\Omega}^{\mathrm{T}}(\mathbf{E}_{ij}\Phi^{\mathrm{T}} + \Phi \mathbf{E}_{ij})\alpha_{\Omega}] + \frac{n}{2}\tr[\Omega^{-1}(\mathbf{E}_{ij}\Phi^{\mathrm{T}} + \Phi \mathbf{E}_{ij})], \\
  \frac{\partial \mathcal{L}}{\partial \varphi_{ii}} & = -\frac{(\tau +d)}{2}\tr[U^{-1}\alpha_{\Omega}^{\mathrm{T}}(\mathbf{J}_{ii}\Phi^{\mathrm{T}} + \Phi \mathbf{J}_{ii})\alpha_{\Omega}] + \frac{n}{2}\tr[\Omega^{-1}(\mathbf{J}_{ii}\Phi^{\mathrm{T}} + \Phi \mathbf{J}_{ii})],
\end{align*}
where $U = K' + Y\Omega^{-1}Y^{\mathrm{T}}$, $\tau = \nu + n -1$ and $\psi_n(\cdot)$ is the derivative of the function $\ln \Gamma_n(\cdot)$ with respect to $\nu$. The details are provided
in \ref{app:MV-TPR-Gradient}.

\begin{Remark}\label{rmk:remark}
It is known that the marginal likelihood function in GPR models is not usually convex with respect to the hyper-parameters, therefore the optimization algorithm may converge to a local optimum
whereas the global one provides better result \cite{cite:autokernel}.
As a result, the optimized hyper-parameters obtained by maximum likelihood estimation
and the performance of GPR models may depend on the initial values of the optimization algorithm \cite{cite:WCKI,cite:BBS,wilson2014covariance,mackay1998introduction}.
A common strategy adopted by most GPR practitioners is a heuristic method. That is, the optimization
is repeated using several initial values generated randomly from a prior distribution based on their expert opinions and experiences, for example, using 10 initial values randomly
selected from a uniform distribution. The final estimates of the hyper-parameters are the ones with the largest likelihood values after
convergence \cite{cite:WCKI,cite:BBS,wilson2014covariance}. Further discussion on
 how to select suitable initial hyper-parameters can be found in \cite{wilson2014covariance,chen2018priors}. In our numerical experiments, 
 the same heuristic method is used for both MV-GPR and MV-TPR.
\end{Remark}

\section{Experiments}\label{sec:experiments}
In this section, we demonstrate the usefulness of MV-GPR and our proposed MV-TPR
using some numerical examples, including simulated data and real data.

\subsection{Simulated Example}
We first use simulation examples to evaluate the quality of the parameter estimation
and the prediction performance of the proposed models.

\subsubsection{Evaluation of parameter estimation}
We generate random samples from a bivariate Gaussian process
 $\bm{y} \sim \mathcal{MGP}(0,k',\Omega)$, where $k'$ is defined as in \eqref{kenerl-with-noise}
 with the kernel $k_{SE}$. The hyper-parameters in $k_{SE}$ are set as
 $[\ell,s_f^2] = [0.5, 2.5]$ and $\Omega = \left( \begin{smallmatrix} 1 & 0.8 \\ 0.8 & 1 \end{smallmatrix} \right)$. The variance of the random noise in $k'$ takes values
 $\sigma_n^2=0.1$, $0.05$ and $0.01$. As explained in Section 3.1, in our models
 the random noises are included in the kernel, therefore
no additional random errors can be added when the random samples are generated;
otherwise two random error terms will result in identifiability issues
in the parameter estimation.
The covariate $x$ has 100 equally spaced values in [0, 1].
 We utilize the heuristic method discussed in the Remark \ref{rmk:remark} to estimate the parameters.
 The experiment is repeated 20 times and we use the parameter Median Relative Error (pMRE) as
 a measure of the quality of the estimates \cite{boukouvalas2014optimal}:
\begin{equation*}
    \mbox{pMRE} = \mbox{median} \left\{ \frac{|\hat{\theta}_i - \theta|}{\theta}, i = 1,2,\cdots 20 \right\},
\end{equation*}
where $|\cdot|$ is the absolute value, $\hat{\theta}_i$ is the parameter estimates
in repetition $i$ and $\theta$ is the true parameter. The results are shown
in Table \ref{tab:MV-GP_parameter_est}.
\begin{table}[htbp]
  \centering
   \caption{The pMREs of MV-GP samples with different noise levels estimated by MV-GPR}
\begin{tabular}{c|ccc}
 \toprule
noise level $\sigma^2_n$  & 0.1 & 0.05 & 0.01 \\
  \midrule
$\hat{\sigma}^2_n$    & 0.884 & 0.931 & 0.974 \\
$\hat{s}_f^2$    & 0.199 & 0.325 & 0.370 \\
$\hat{\ell}^2$   & 0.061 & 0.050 & 0.034 \\
$\hat{\varphi}_{11}$  & 1.619 & 1.266 & 0.931 \\
$\hat{\varphi}_{22}$  & 1.586 & 1.372 & 0.901 \\
$\hat{\phi}_{12}$  & 0.846 & 0.878 & 0.858 \\
\bottomrule
\end{tabular}
  \label{tab:MV-GP_parameter_est}%
\end{table}

\begin{table}[htbp]
  \centering
   \caption{The pMREs of MV-TP samples with different noise levels estimated by MV-TPR}
\begin{tabular}{c|ccc}
 \toprule
noise level $\sigma^2_n$  & 0.1 & 0.05 & 0.01 \\
  \midrule
$\hat{\sigma}^2_n$    & 0.756       & 0.714        & 0.758        \\
$\hat{s}_f^2$    & 0.522       & 0.429        & 1.283        \\
$\hat{\ell}^2$    & 0.120       & 0.165        & 0.097        \\
$\hat{\varphi}_{11}$ & 1.167       & 1.079        & 0.898        \\
$\hat{\varphi}_{22}$ & 1.174       & 1.078        & 0.903        \\
$\hat{\phi}_{12}$  & 0.951       & 0.981        & 0.965        \\
$\hat{\nu}$          & 0.192       & 0.247        & 0.245        \\
\bottomrule
\end{tabular}
  \label{tab:MV-TP_parameter_est}%
\end{table}

The similar experiment is also conducted for MV-TPR, where the samples are generated
from a bivariate Student$-t$ process $\bm{y} \sim \mathcal{MTP}(\nu,0, k',\Omega)$
with the same true parameters as above and $\nu = 3$.
The results are reported in Table \ref{tab:MV-TP_parameter_est}.

It can be seen that the length scale ${\ell}^2$ is well estimated in both cases,
while the estimates for the parameters in the row covariance matrix
(${\varphi}_{11}$, ${\varphi}_{22}$ and ${\phi}_{12}$)
are not as good but reasonable. This may be because the conjugate gradient optimization
algorithm does not manage to reach the global maxima of the likelihood function.
Better results may be achieved using optimal design for parameter
estimation as discussed in \cite{boukouvalas2014optimal}
In general the estimates of the parameters are closer
to the true values as the noise level decreases, but the improvement
in terms of estimation accuracy is not significant.

\subsection{Evaluation of prediction accuracy}
Now we consider a simulated data from two specific functions. The true model used to generate data is given by
\begin{eqnarray*}
  \bm{y} &=& [f_1(x), f_2(x)] + [\varepsilon^{(1)} , \varepsilon^{(2)}],   \\
  f_1(x) &=&  2x \cos(x), \quad f_2(x) = 1.5x \cos(x + \pi/5) ,
\end{eqnarray*}
where the random noise $[\varepsilon^{(1)} , \varepsilon^{(2)}] \sim  \mathcal{MGP}(0,k_{SE},\Omega)$.
We select $k_{SE}$ with parameter $[\ell,s_f^2] = [1.001,5]$ and $\Omega = \left( \begin{smallmatrix} 1 & 0.25 \\ 0.25 & 1 \end{smallmatrix} \right)$. The covariate $x$ has 100 equally spaced values in [-10, 10] so that a sample of 100 observations for $y_1$ and $y_2$ are obtained.

\begin{figure}[htbp]
  \centering
  \subfigure[MV-GPR ($y_1$)]{
     \raisebox{-2cm}{\includegraphics[width=0.20\textwidth]{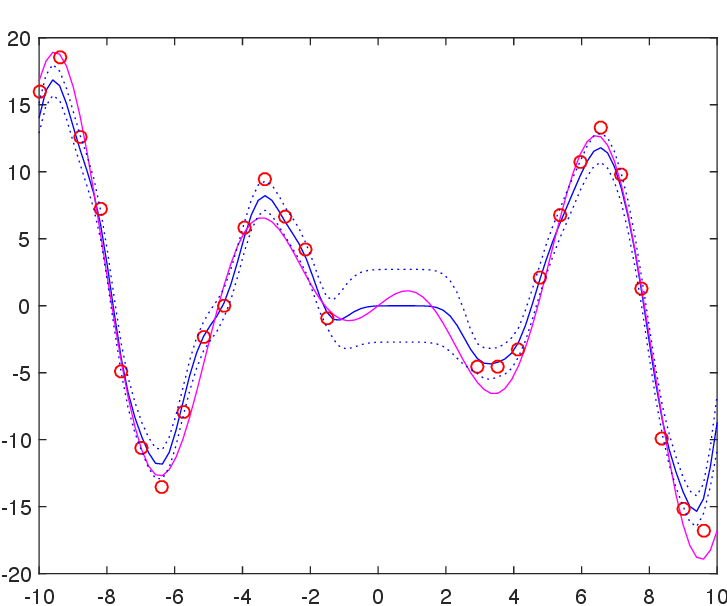}} }
  \subfigure[MV-TPR ($y_1$)]{
     \raisebox{-2cm}{\includegraphics[width=0.20\textwidth]{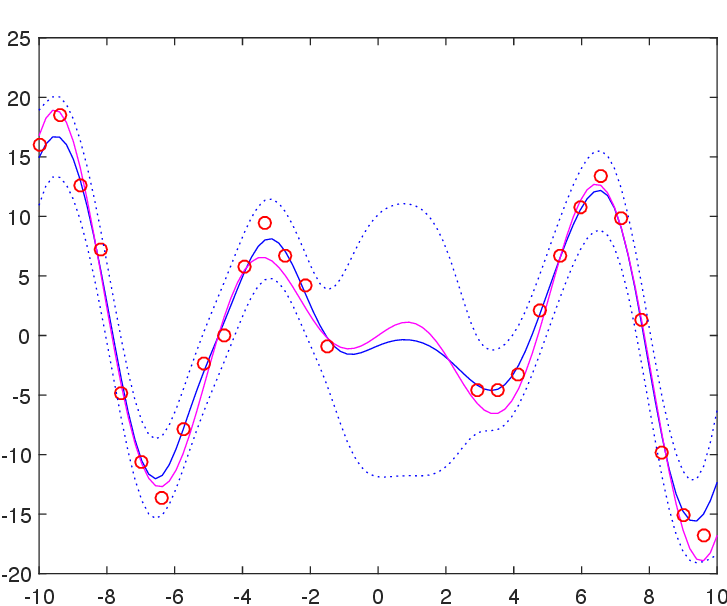}} }
  \subfigure[GPR ($y_1$)]{
     \raisebox{-2cm}{\includegraphics[width=0.20\textwidth]{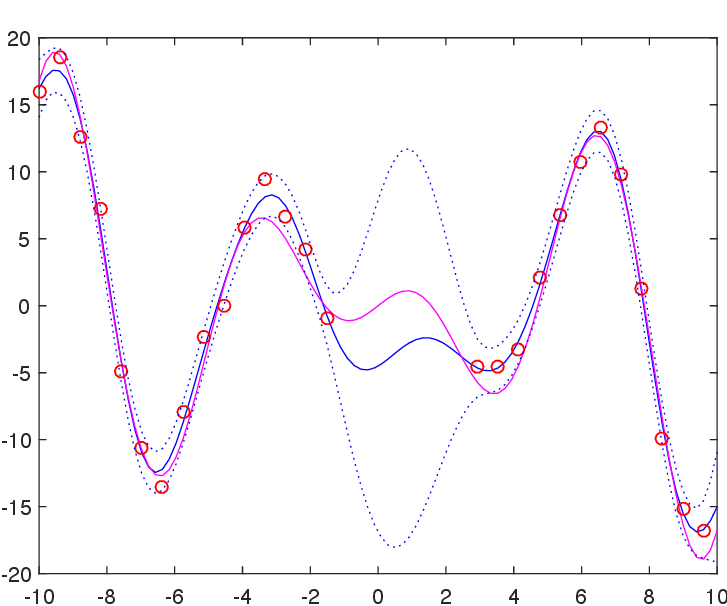}} }
  \subfigure[TPR ($y_1$)]{
     \raisebox{-2cm}{\includegraphics[width=0.20\textwidth]{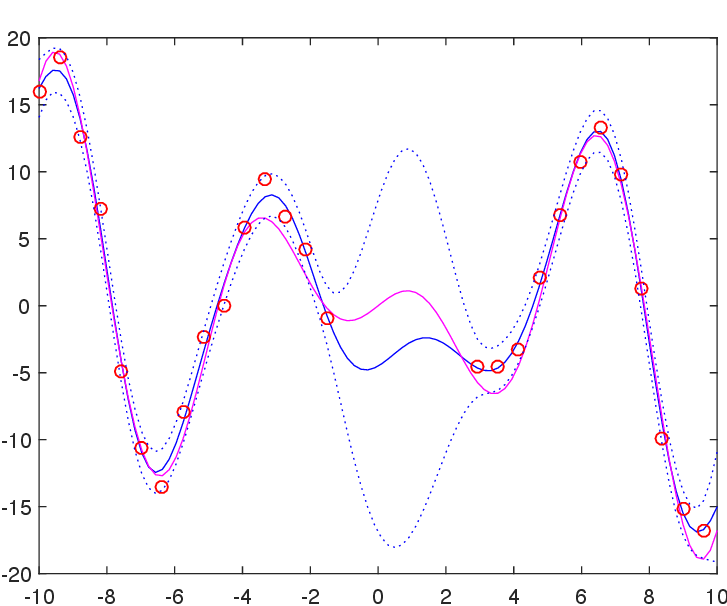}} }
  \subfigure[MV-GPR ($y_2$)]{
     \raisebox{-2cm}{\includegraphics[width=0.20\textwidth]{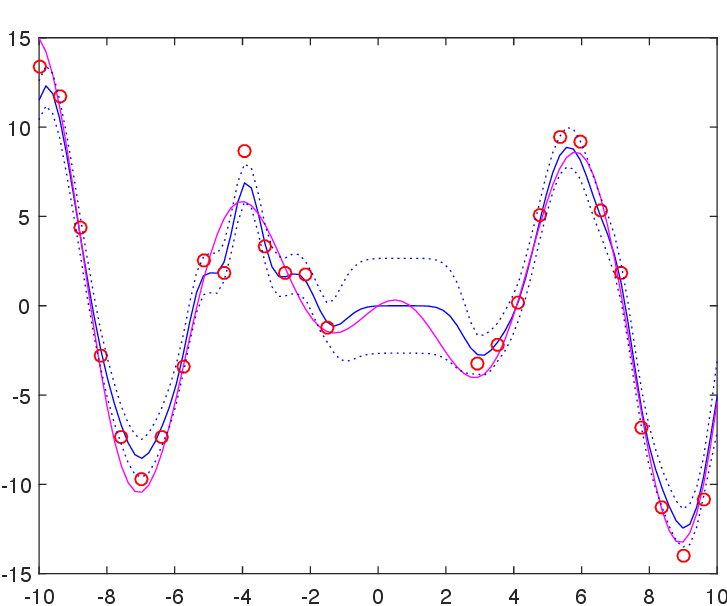}} }
  \subfigure[MV-TPR ($y_2$)]{
     \raisebox{-2cm}{\includegraphics[width=0.20\textwidth]{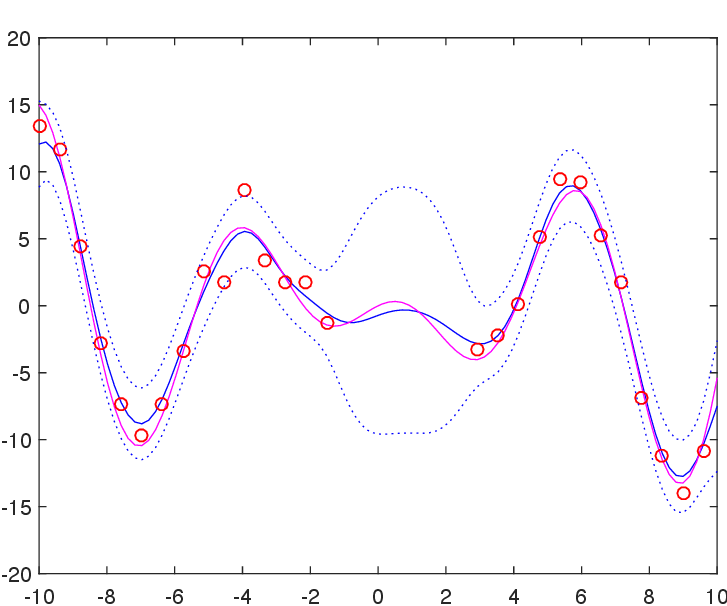}} }
  \subfigure[GPR ($y_2$)]{
     \raisebox{-2cm}{\includegraphics[width=0.20\textwidth]{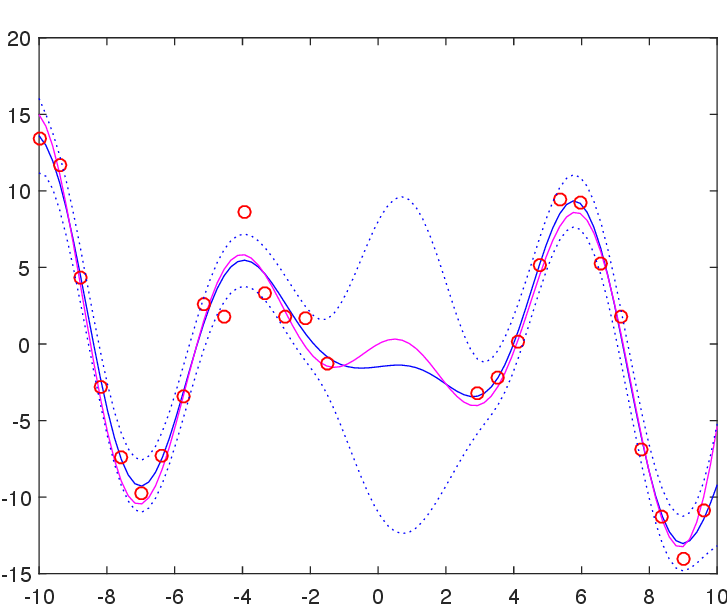}} }
  \subfigure[TPR ($y_2$)]{
     \raisebox{-2cm}{\includegraphics[width=0.20\textwidth]{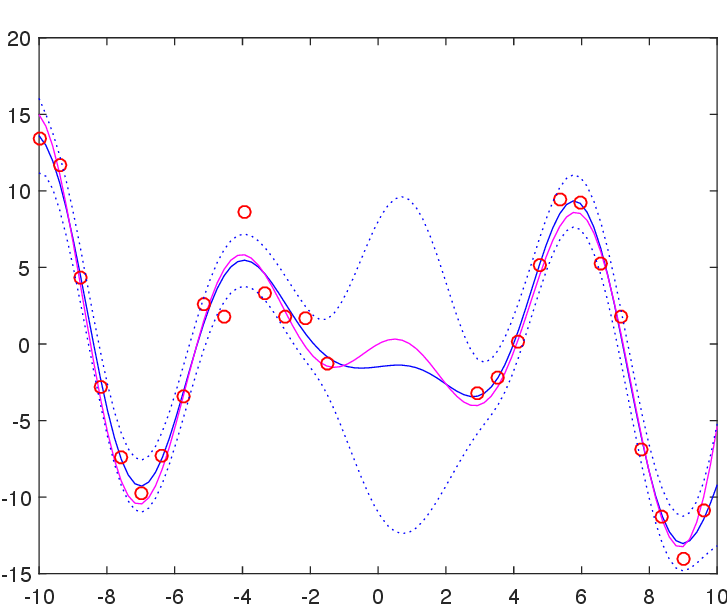}} }
  \caption{Predictions for MV-GP noise data using different models. From panels (\textbf{a}) to (\textbf{d}): predictions for $y_1$ by MV-GPR, MV-TPR, GPR and TPR. From panels (\textbf{e}) to (\textbf{h}): predictions for $y_2$ by MV-GPR, MV-TPR, GPR and TPR. The solid blue lines are predictions, the solid red lines are the true functions and the circles are the observations. The dash lines represent the 95\% confidence intervals}\label{fig:MV-gptpG}
\end{figure}

For model training, we use fewer points with one part missing so that the $z$th
training data points with $z = \{3r+1\}_{r=1}^{12} \cup \{3r+2\}_{r=22}^{32}$ are selected for both $y_1$ and $y_2$. The prediction is then performed for
 all 100 covariate values in [-10, 10]. The RMSEs between the predicted values and the true ones from $f_1(x)$ and $f_2(x)$ are calculated.
 For comparison, the conventional GPR and TPR models are also applied
 to the two outputs independently.
 The above experiment is repeated 1000 times and the ARMSE
 (Average Root Mean Square Error), defined by
$$
\text{ARMSE} = \frac{1}{1000}\sum_{i=1}^{1000}\left(\frac{1}{100}\sum_{j=1}^{100}(\hat{y}_{ij} - y_{ij})^2\right)^{\frac{1}{2}},
$$
is calculated. Here $y_{ij}$ is the $j$th observation in the $i$th
repetition and $\hat{y}_{ij}$ is the corresponding predicted value.
The results are reported in Table \ref{tab:RMSEgpNoise} and an example of prediction is demonstrated in Figure \ref{fig:MV-gptpG}.

\begin{table}[htbp]
  \setlength{\abovecaptionskip}{0pt}
  \setlength{\belowcaptionskip}{5pt}
  \centering
  \caption{The ARMSE by the different models (multivariate Gaussian noisy data)}
    \begin{tabular}{cccc|ccccc}
    \toprule
    \multicolumn{4}{c|}{\textbf{Output 1} ($y_1$)}        & \multicolumn{1}{c}{} & \multicolumn{4}{c}{\textbf{Output 2} ($y_2$)} \\
    \midrule
     \multicolumn{1}{c}{MV-GPR} & \multicolumn{1}{c}{MV-TPR} & \multicolumn{1}{c}{GPR} & \multicolumn{1}{c|}{TPR} & \multicolumn{1}{c}{} & \multicolumn{1}{c}{MV-GPR} & \multicolumn{1}{c}{MV-TPR} & \multicolumn{1}{c}{GPR} & \multicolumn{1}{c}{TPR} \\
    \midrule
    1.540 & \textbf{1.258} & 1.594 & 1.585 &       & 1.749 & \textbf{1.518} &  2.018& 2.017 \\
    \bottomrule
    \end{tabular}%
  \label{tab:RMSEgpNoise}%
\end{table}%

\begin{figure}[htbp]
  \centering
  \subfigure[MV-GPR ($y_1$)]{
     \raisebox{-2cm}{\includegraphics[width=0.20\textwidth]{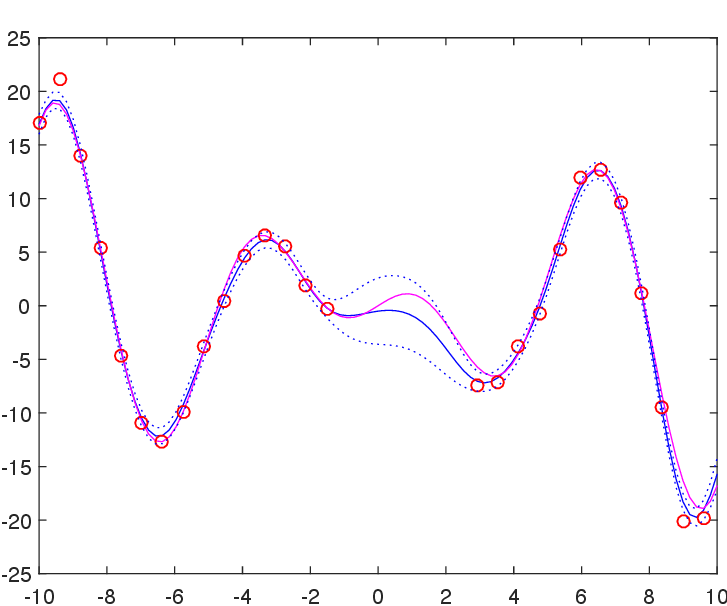}} }
  \subfigure[MV-TPR ($y_1$)]{
     \raisebox{-2cm}{\includegraphics[width=0.20\textwidth]{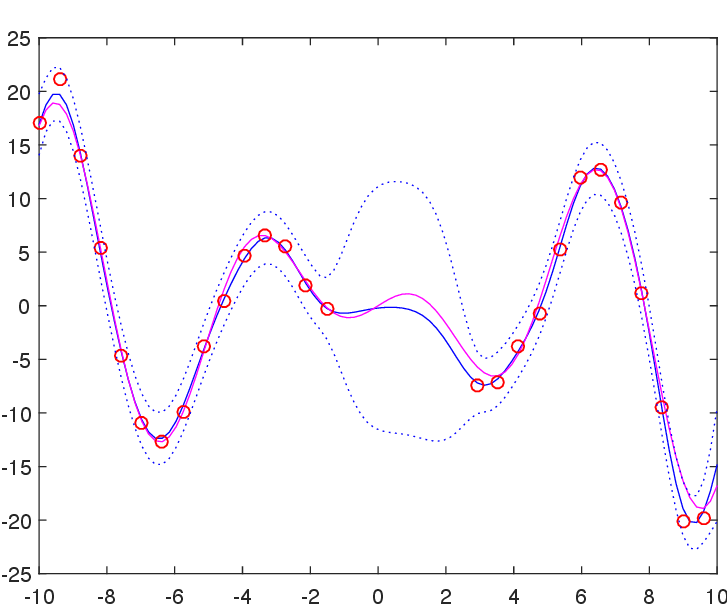}} }
  \subfigure[GPR ($y_1$)]{
     \raisebox{-2cm}{\includegraphics[width=0.20\textwidth]{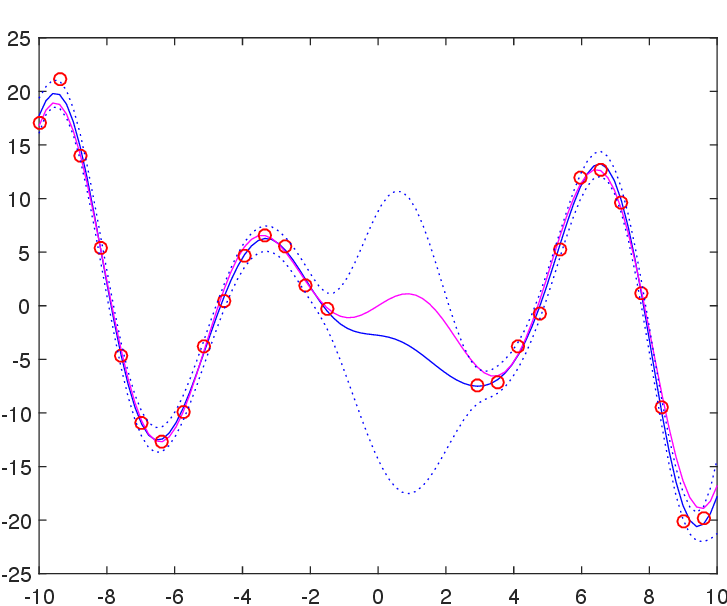}} }
  \subfigure[TPR ($y_1$)]{
     \raisebox{-2cm}{\includegraphics[width=0.20\textwidth]{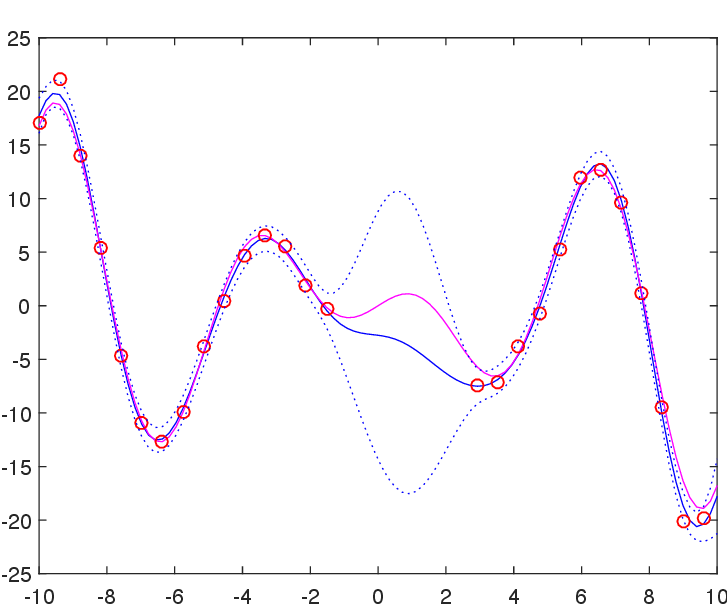}} }
  \subfigure[MV-GPR ($y_2$)]{
     \raisebox{-2cm}{\includegraphics[width=0.20\textwidth]{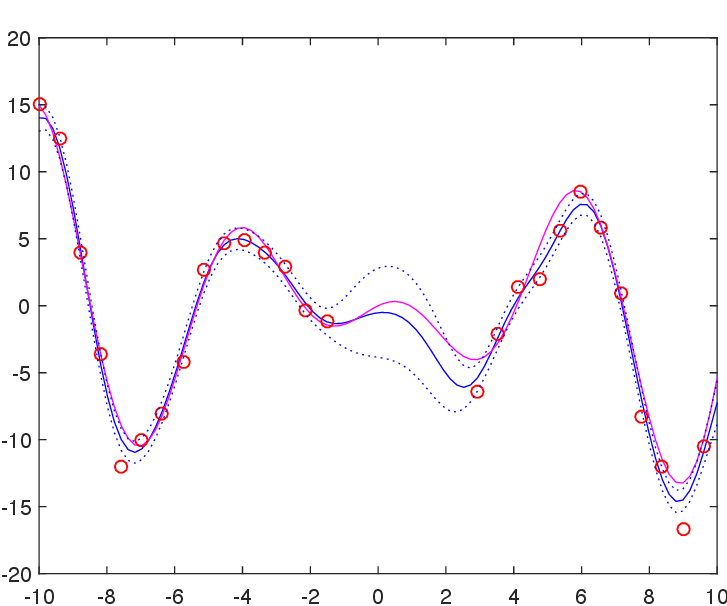}} }
  \subfigure[MV-TPR ($y_2$)]{
     \raisebox{-2cm}{\includegraphics[width=0.20\textwidth]{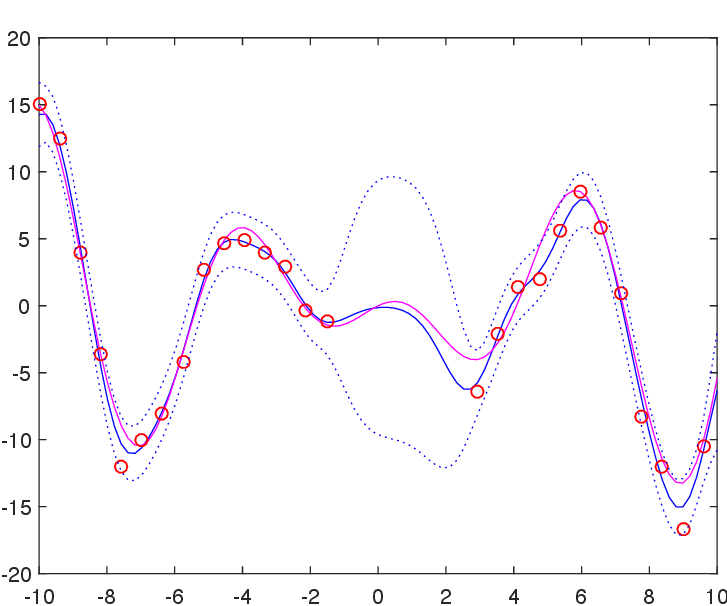}} }
  \subfigure[GPR ($y_2$)]{
     \raisebox{-2cm}{\includegraphics[width=0.20\textwidth]{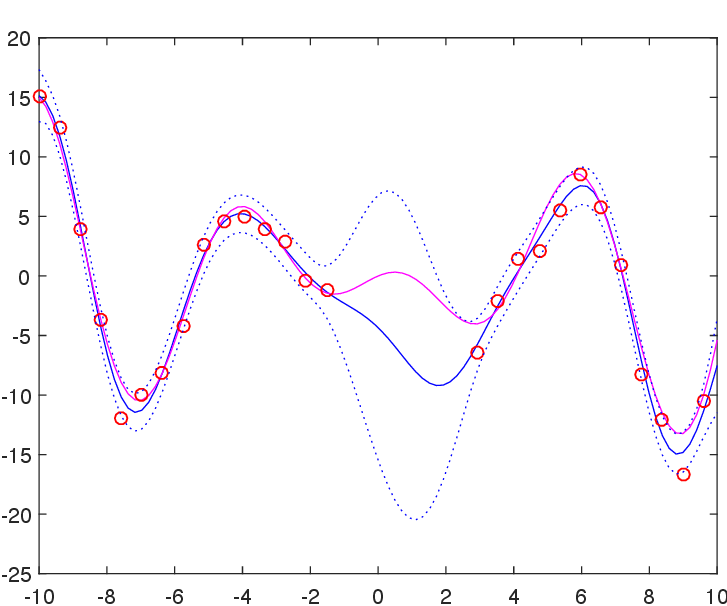}} }
  \subfigure[TPR ($y_2$)]{
     \raisebox{-2cm}{\includegraphics[width=0.20\textwidth]{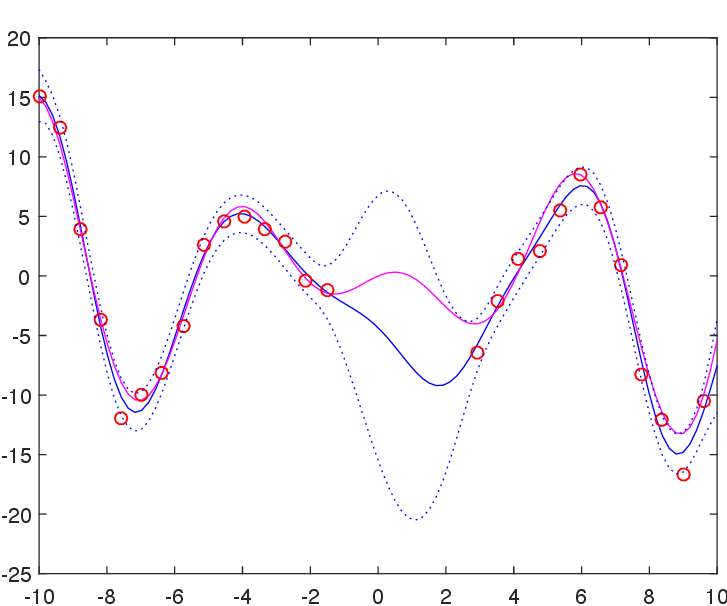}} }
  \caption{Predictions for MV-TP noise data using different models. From panels (\textbf{a}) to (\textbf{d}): predictions for $y_1$ by MV-GPR, MV-TPR, GPR and TPR. From panels (\textbf{e}) to (\textbf{h}): predictions for $y_2$ by MV-GPR, MV-TPR, GPR and TPR. The solid blue lines are predictions, the solid red lines are the true functions and the circles are the observations. The dash lines represent the 95\% confidence intervals}\label{fig:MV-gptpT}
\end{figure}

The similar experiment is also conducted for the case where
the random noise follows a multivariate Student$-t$ process
$[\varepsilon^{(1)} , \varepsilon^{(2)}] \sim \mathcal{MTP}(3,0,k_{SE},\Omega)$.
We select $k_{SE}$ with parameter $[\ell,s_f^2] = [1.001,5]$
and $\Omega = \left( \begin{smallmatrix} 1 & 0.25 \\ 0.25 & 1 \end{smallmatrix} \right)$.
The resulted ARMSEs are presented in Table \ref{tab:RMSEtpNoise} and
an example of prediction is demonstrated in Figure \ref{fig:MV-gptpT}.
\begin{table}[htbp]
  \setlength{\abovecaptionskip}{0pt}
  \setlength{\belowcaptionskip}{5pt}
  \centering
  \caption{The ARMSE by the different models (multivariate Student$-t$ noisy data)}
    \begin{tabular}{cccc|ccccc}
    \toprule
     \multicolumn{4}{c|}{\textbf{Output 1} ($y_1$)}        & \multicolumn{1}{c}{} & \multicolumn{4}{c}{\textbf{Output 2} ($y_2$)} \\
    \midrule
     \multicolumn{1}{c}{MV-GPR} & \multicolumn{1}{c}{MV-TPR} & \multicolumn{1}{c}{GPR} & \multicolumn{1}{c|}{TPR} & \multicolumn{1}{c}{} & \multicolumn{1}{c}{MV-GPR} & \multicolumn{1}{c}{MV-TPR} & \multicolumn{1}{c}{GPR} & \multicolumn{1}{c}{TPR} \\
    \midrule
    1.441 & \textbf{1.238} & 1.505 & 1.503 &       & 1.636 & \textbf{1.464} &  1.941& 1.940 \\
    \bottomrule
    \end{tabular}%
  \label{tab:RMSEtpNoise}%
\end{table}%

From the tables and figures above, it can be seen that the multivariate process regression models are able to discover a more desired pattern in the gap compared with the conventional GPR and TPR models used independently. It also reveals that taking the correlations between the two outputs into consideration improves the accuracy of prediction compared with the methods of modelling each output independently. In particular, MV-TPR performs better than MV-GPR in the predictions
for both types of noisy data whilst the performances by TPR and GPR are similar.

It is not surprising that
in general MV-TPR works better than MV-GPR when the outputs have dependencies,
 because the former has more modelling flexibility with one more parameter which captures
  the degree of freedom. Theoretically, MV-TP converges to MV-GP if the degree of freedom tends to infinity, and to some extent MV-GPR is a special case of MV-TPR. In the above experiment
because the observations, even generated from MV-GP, may contain outliers and the sample size
(23 training points) is small, MV-TPR with a large degree of freedom may fit the data better than MV-GPR and
hence provides a better prediction. These findings coincide with those in
 \cite{shah2014student} in the context of univariate Student$-t$ process regressions.
 On the other hand, while a training sample of size 23 may be too small for two-dimensional processes such that MV-TPR performs better than MV-GPR, it may be large enough for one-dimensional processes, leading to similar performances by TPR and GPR. 

It is noted that the predictive variance by MV-GPR is much smaller than the independent GPR model. This is likely due to the loss of information in the independent model.

\subsection{Real data examples}
We further test our proposed methods on two real datasets \footnote{These data sets are from the UC Irvine Machine Learning Repository (\url{https://archive.ics.uci.edu/ml/index.php}).}. The selected mean function is zero-offset and the selected kernel is SEard. Before
the experiments are conducted, all the data have been normalized by
$$
\tilde{y}_i = \frac{y_i-\mu}{\sigma},
$$
where $\mu$ and $\sigma$ are the sample mean and standard deviation of the data $\{y_i\}_{i=1}^n$ respectively.

\subsubsection{Bike rent prediction}
This dataset contains the hourly and daily count of rental bikes between years 2011 and 2012 in Capital Bikeshare System with the corresponding weather and seasonal information \cite{fanaee2014event}. There are 16 attributes. We test our proposed methods for multi-output prediction based on daily count dataset. After deleting all the points with missing attributes, we use the first 168 data points in the season Autumn because the data is observed on a daily basis (1 week = 7 days) and the whole dataset is divided into 8 subsets (each subset has 3 weeks' data points). In the experiment, the input comprises 8 attributes, including normalized temperature, normalized feeling temperature, normalized humidity, normalized wind speed, whether the day is holiday or not, day of the week, working day or not and weathersit.
The output consists of 2 attributes, including the count of casual users (Casual) and the count of registered users (Registered).

The cross-validation method is taken as $k-$fold, where $k = 8$.
Each subset is considered as a test set and the remaining subsets are considered as a training set. Four models, including MV-GPR, MV-TPR, GPR (to predict each output independently) and TPR (to predict each output independently) are applied to the data and predictions are made based on the divided
training and test sets. The process is repeated for 8 times, and
for each subset's prediction, the MSE (mean square error) and the MAE (mean absolute error) is calculated. The medians of the 8 MSEs and MAEs are then used to evaluate the performance
for each output. Finally, the maximum median of all the outputs (MMO) is used to evaluate
the overall performance of the multi-dimensional prediction.
The results are shown in Table \ref{tab:BikeDetails}.
It can be seen that MV-TPR significantly outperforms all the other models
in terms of MSE and MAE, and MV-GPR performs the second best, whilst
TPR is slightly better than GPR. 

\begin{table}[htbp]
 \caption{Bike rent prediction results based on MSEs and MAEs}
 \centering
  \subtable[MSE]{
    \begin{tabular}{c|c|cccc}
    \toprule
          &       & \multicolumn{1}{c}{\textbf{MV-GPR}} & \multicolumn{1}{c}{\textbf{MV-TPR}} & \multicolumn{1}{c}{\textbf{GPR}} & \multicolumn{1}{c}{\textbf{TPR}} \\
    \midrule
    \multicolumn{1}{c|}{Outputs (median} & Casual & 0.411 & \textbf{0.334} & 0.424 & 0.397 \\
    \multicolumn{1}{c|}{ of 8 subsets' MSEs)} & Registered & 0.982 & \textbf{0.903} & 1.134 & 1.111 \\
    \midrule
    \multicolumn{1}{c|}{MMO} &       & 0.982 & \textbf{0.903} & 1.134 & 1.111 \\
    \bottomrule
    \end{tabular}%
        \label{tab:MSEbike}
 }
 \subtable[MAE]{
    \begin{tabular}{c|c|cccc}
    \toprule
          &       & \multicolumn{1}{c}{\textbf{MV-GPR}} & \multicolumn{1}{c}{\textbf{MV-TPR}} & \multicolumn{1}{c}{\textbf{GPR}} & \multicolumn{1}{c}{\textbf{TPR}} \\
    \midrule
     \multicolumn{1}{c|}{Outputs (median} & Casual & 0.558 & \textbf{0.488} & 0.540 & 0.546 \\
    \multicolumn{1}{c|}{ of 8 subsets' MAEs)} & Registered & 0.897 & \textbf{0.855} & 0.916 & 0.907 \\
    \midrule
    \multicolumn{1}{c|}{MMO} &       & 0.897 & \textbf{0.855} & 0.916 & 0.907 \\
    \bottomrule
    \end{tabular}%
        \label{tab:MAEbike}
 }
  \label{tab:BikeDetails}
\end{table}

\subsubsection{Air quality prediction}
The dataset contains 9,358 instances of hourly averaged responses from an array of 5 metal oxide chemical sensors embedded in an Air Quality Chemical Multisensor Device with 15 attributes \cite{de2008field}. We delete all the points with missing attributes (887 points remaining). The first 864 points are considered in our experiment because the data is hourly observed (1 day = 24 hours) and the whole data set is divided into 9 subsets (each subset has 4-days' data points, totally 864 data points). In the experiment, the input comprises 9 attributes, including time, true hourly averaged concentration CO in $mg/m^3$ (COGT), true hourly averaged overall Non Metanic HydroCarbons concentration in $microg/m^3$ (NMHCGT), true hourly averaged Benzene concentration in $microg/m^3$ (C6H6GT), true hourly averaged NOx concentration in $ppb$ (NOx), true hourly averaged NO2 concentration in $microg/m^3$ (NO2), absolute humidity (AH), temperature (T) and relative humidity (RH). The output consists of 5 attributes, including PT08.S1 (tin oxide) hourly averaged sensor response, PT08.S2 (titania) hourly averaged sensor response, PT08.S3 (tungsten oxide) hourly averaged sensor response, PT08.S4 (tungsten oxide) hourly averaged sensor response and PT08.S5 (indium oxide) hourly averaged sensor response.

The cross-validation method is taken as $k-$fold, where $k = 9$.
The remaining modelling
procedure is the same as in the bike rent prediction experiment,
except that $k$ is 9 so that the process is repeated 9 times.
The results are shown in Table \ref{tab:AirDetails}.
It can be observed that MV-TPR consistently outperforms MV-GPR, and in overall terms, MV-GPR does not perform as well as independent GPR.
These results can likely be explained as follows.
In fact, in our proposed framework, all the outputs are considered as a whole and the same kernel (and hyperparameters) is used for all the outputs, which may not be appropriate if different outputs have very different patterns (which is the case in this example).
On the other hand, in the independent modelling different GPR models are used for different outputs and they can have different hyperparameters even though the kernel is the same, which may result in better prediction accuracy. It is noted that in this example
MV-TPR still works better than MV-GPR because the former offers more modelling flexibility and is thus more robust to model misspecification. This finding is consistent with that by \cite{shah2014student} for the univariate Student$-t$ process.

\begin{table}[htbp]
 \caption{Air quality prediction results based on MSEs and MAEs}
 \centering
  \subtable[MSE]{
    \begin{tabular}{c|c|cccc}
    \toprule
          &       & \multicolumn{1}{c}{\textbf{MV-GPR}} & \multicolumn{1}{c}{\textbf{MV-TPR}} & \multicolumn{1}{c}{\textbf{GPR}} & \multicolumn{1}{c}{\textbf{TPR}} \\
    \midrule
          & PT08S1CO & 0.091 & 0.065 & 0.079 & 0.074 \\
    \multicolumn{1}{c|}{Outputs} & PT08S2NMHC & $8.16\times 10^{-5}$ & $3.42\times 10^{-5}$ & $1.91\times 10^{-7}$ & $7.32\times 10^{-8}$ \\
    \multicolumn{1}{c|}{(Median of 9} & PT08S3NOx & 0.036 & 0.027 & 0.022 & 0.025 \\
    \multicolumn{1}{c|}{ subsets' MSEs)} & PT08S4NO2 & 0.015 & 0.014 & 0.010 & 0.009 \\
    \multicolumn{1}{c|}{} & PT08S5O3 & 0.092 & 0.073 & 0.060 & 0.067 \\
        \midrule
    \multicolumn{1}{c|}{MMO} &       & 0.092 & \textbf{0.073} & 0.079 & 0.074 \\
    \bottomrule
    \end{tabular}%
        \label{tab:MSE}
 }
 \subtable[MAE]{
    \begin{tabular}{c|c|cccc}
    \toprule
          &       & \multicolumn{1}{c}{\textbf{MV-GPR}} & \multicolumn{1}{c}{\textbf{MV-TPR}} & \multicolumn{1}{c}{\textbf{GPR}} & \multicolumn{1}{c}{\textbf{TPR}} \\
    \midrule
          & PT08S1CO & 0.240 & 0.204 & 0.212 & 0.223 \\
    \multicolumn{1}{c|}{Outputs} & PT08S2NMHC & $6.39\times 10^{-3}$ & $1.15\times 10^{-2}$ & $1.80\times 10^{-4}$ & $9.26\times 10^{-5}$ \\
    \multicolumn{1}{c|}{(Median of 9} & PT08S3NOx & 0.141 & 0.122 & 0.115 & 0.120 \\
    \multicolumn{1}{c|}{ subsets' MAEs)} & PT08S4NO2 & 0.095 & 0.089 & 0.079 & 0.073 \\
    \multicolumn{1}{c|}{} & PT08S5O3 & 0.231 & 0.210 & 0.199 & 0.205 \\
    \midrule
    \multicolumn{1}{c|}{MMO} &   & 0.240 & \textbf{0.210} & 0.212 & 0.223 \\
    \bottomrule
    \end{tabular}%
        \label{tab:MAE}
 }
  \label{tab:AirDetails}
\end{table}

\subsection{Application to stock market investment}

In the previous subsections, the examples show the usefulness of our proposed methods in terms of more accurate prediction. Furthermore, our proposed methods can be applied to produce trading strategies in the stock market investment.

It is known that the accurate prediction of future for an equity market is almost impossible. Admittedly, the more realistic idea is to make a strategy based on the Buy\&Sell signal in the different prediction models \cite{akbilgic2014novel}. In this paper, we consider a developed Dollar 100 (dD100) as a criterion of the prediction models. The dD100 criterion is able to reflect the theoretical future value of \$100 invested at the beginning, and traded according to the signals constructed by predicted value and the reality. The details of dD100 criterion are described in Section \ref{Section:Strategy}.

Furthermore, the equity index is an important measurement of the value of a stock market and is used by many investors making trades and scholars studying stock markets. The index is computed from the weighted average of the selected stocks’ prices, so it is able to describe how the whole stock market in the consideration performs in a period and thus many trading strategies of a stock or a portfolio have to take the information of the index into account. As a result, our experimental predictions for specific stocks are based on the indices as well.

\subsubsection{Data preparation}
We obtain daily price data, containing opening, closing, and adjusted closing for the stocks (the details are shown in Section \ref{Section:Chinese} and Section \ref{Section:Dow30}) and three main indices in the US, Dow Jones Industrial Average (INDU), S\&P500 (SPX), and NASDAQ (NDX) from Yahoo Finance in the period of 2013 -- 2014. The log returns of the adjusted closing price and inter-day log returns are obtained by defining
\begin{align*}
  &\text{Log return: } LR_i = \ln \frac{ACP_i}{ACP_{i-1}}, \\
  &\text{Inter-day log return: } ILR_i = \ln \frac{CP_i}{OP_i},
\end{align*}
where $ACP_i$ is the adjusted closing price of the $i$th day ($i>1$), $CP_i$ is the closing price of the $i$th day, and $OP_i$ is the opening price of the $i$th day. Therefore, there are totally 503 daily log returns and log inter-day returns for all the stocks and indices from 2013 to 2014.

\subsubsection{Prediction model and strategy}\label{Section:Strategy}

The sliding windows method is used for our prediction models, including GPR, TPR, MV-GPR, and MV-TPR, based on the indices, INDU, SPX, and NDX. The training sample is set as 303, which is used to forecast for the next 10 days, and the training set is updated by dropping off the earliest 10 days and adding on the latest 10 days when the window is moved. The sliding-forward process was run 20 times, resulting in a total of 200 prediction days, in groups of 10. The updated training set allows all the models and parameters to adapt the dynamic structure of the equity market \cite{akbilgic2014novel}. Specifically, the inputs consist of the log returns of 3 indices, the targets are multiple stocks' log returns and Standard Exponential with automatic relevance determination (SEard) is used as the kernel for all of these prediction models.

It is noteworthy that the predicted log returns of stocks are used to produce a buy or sell signal for trading rather than to discover an exact pattern of the future. The signal $BS$ produced by the predicted log returns of the stocks is defined by
$$
BS_i = \hat{LR}_i - LR_i + ILR_i, i = 1,\cdots ,200,
$$
where $\{\hat{LR}_i\}_{i=1}^{200}$ are the predicted log returns of a specific stock, $\{LR_i\}_{i=1}^{200}$ are the true log returns while $\{ILR_i\}_{i=1}^{200}$ are the inter-day log returns. The Buy\&Sell strategy relying on the signal $BS$ is described in Table \ref{tab:iD100}.
\begin{table}[H]
  \setlength{\abovecaptionskip}{0pt}
  \setlength{\belowcaptionskip}{5pt}
  \centering
  \caption{Buy\&Sell strategy of dD100 investment}
    \begin{tabular}{cl}
    \toprule
    \textbf{Decision} & \textbf{Condition} \\
    \midrule
    Buy   & $\hat{LR}_i>0$, \& $BS_i>0 $ \& we have the position of cash \\
    Sell  & $\hat{LR}_i<0$, \& $BS_i<0 $ \& we have the position of share \\
    Keep  & No action is taken for the rest of the option \\
    \bottomrule
    \end{tabular}%
  \label{tab:iD100}%
\end{table}%

It is noted that the stocks in our experiment are counted in Dollar rather than the number of shares, which means in theory we can precisely buy or sell a specific Dollar valued stock. For example, if the stock price is \$37 when we only have \$20, we can still buy \$20 valued stock rather than borrowing \$17 and then buy 1 share. Furthermore, it is also necessary to explain why we choose the signal $BS$. By the definition, we rewrite it as
\begin{eqnarray*}
  BS_i  &=& \ln(\frac{\hat{ACP}_i}{ACP_{i-1}}) - \ln(\frac{ACP_i}{ACP_{i-1}}) + \ln(\frac{CP_i}{OP_i}) \\
   &=& \ln(\frac{\hat{ACP}_i}{ACP_{i-1}}) - \ln(\frac{ACP_i}{ACP_{i-1}}) + \ln(\frac{ACP_i}{AOP_i}) \\
   &=& \ln(\frac{\hat{ACP}_i}{AOP_i}) ,
\end{eqnarray*}
where $\{ACP_i\}_{i=0}^{200}$ are the last 201 adjusted closing prices for a stock, $\{CP\}_{i=1}^{200}$ are the last 200 closing prices, and $\{AOP\}_{i=1}^{200}$ are the adjusted opening prices. If $BS_i > 0$, the predicted closing price should be higher than the adjusted opening price, which means we can obtain the inter-day profit by buying the shares at the opening price \footnote{Actually, the value has to be considered as adjusted opening price since all the shares are counted in Dollar. The adjusted opening price is also easy to compute based on the real opening price and the dividend information} as long as the signal based on our predictions is accurate. Meanwhile, the opposite manipulation based on BS strategy means that we can avoid the inter-day loss by selling decisively at the opening price. Furthermore, the reasonable transaction fee 0.025\% is considered in the experiment since the strategy might trade frequently \footnote{ The figure 0.025\% is comprehensive consideration referred to NASDAQ website:\url{http://nasdaq.cchwallstreet.com/}}. As a result, this is a reasonable strategy since we can definitely obtain a profit by buying the shares and cut the loss by selling the shares in time only if our prediction has no serious problem. It is also an executable strategy because the decision is made based on the next day's reality and our prediction models.

At last, $BS$ signal varies in different prediction models so that we denote these Buy\&Sell strategies based on MV-GPR, MV-TPR, GPR, and TPR model as MV-GPR strategy, MV-TPR strategy, GPR strategy, and TPR strategy, respectively.

\subsubsection{Chinese companies in NASDAQ}\label{Section:Chinese}
In recent years, the "Chinese concepts stock" has received an extensive attention among international investors owing to the fast development of Chinese economy and an increasing number of Chinese firms have been traded in the international stock markets \cite{luo2012overseas}. The "Chinese concepts stock" refers to the stock issued by firms whose asset or earning have essential activities in Mainland China. Undoubtedly, all these "Chinese concept stocks" are heavily influenced by the political and economic environment of China together. For this reason, all these stocks have the potential and unneglectable correlation theoretically, which is probably reflected in the movement of stock prices. The performance of multiple targets prediction, which takes the potential relationship into consideration, should be better. Therefore, the first real data example is based on 3 biggest Chinese companies described in Table \ref{tab:matrixstocks}.
\begin{table}[htbp]
  \setlength{\abovecaptionskip}{0pt}
  \setlength{\belowcaptionskip}{5pt}
  \centering
  \caption{Three biggest "Chinese concept" stocks}
    \begin{tabular}{ccl}
    \toprule
    \textbf{Ticker} & \textbf{Exchange} & \textbf{Company} \\
    \midrule
    BIDU   & NASDAQ & Baidu, Inc.  \\
    CTRP   & NASDAQ &Ctrip.com International, Ltd.    \\
    NTES   & NASDAQ &NetEase, Inc.  \\
    \bottomrule
    \end{tabular}%
  \label{tab:matrixstocks}%
\end{table}%

\begin{figure}[htbp]
\centering
\includegraphics[width=0.48\textwidth]{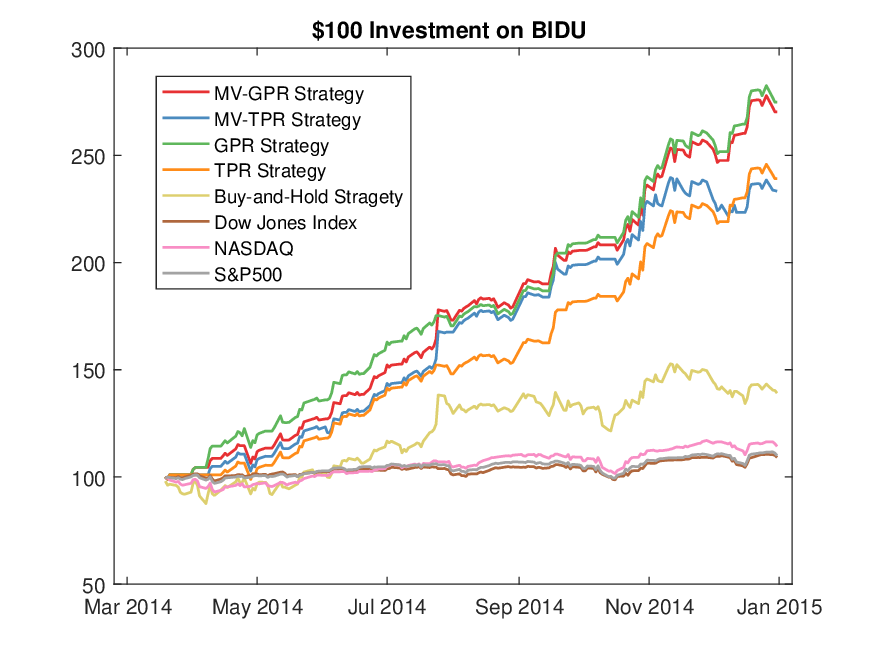}
\includegraphics[width=0.48\textwidth]{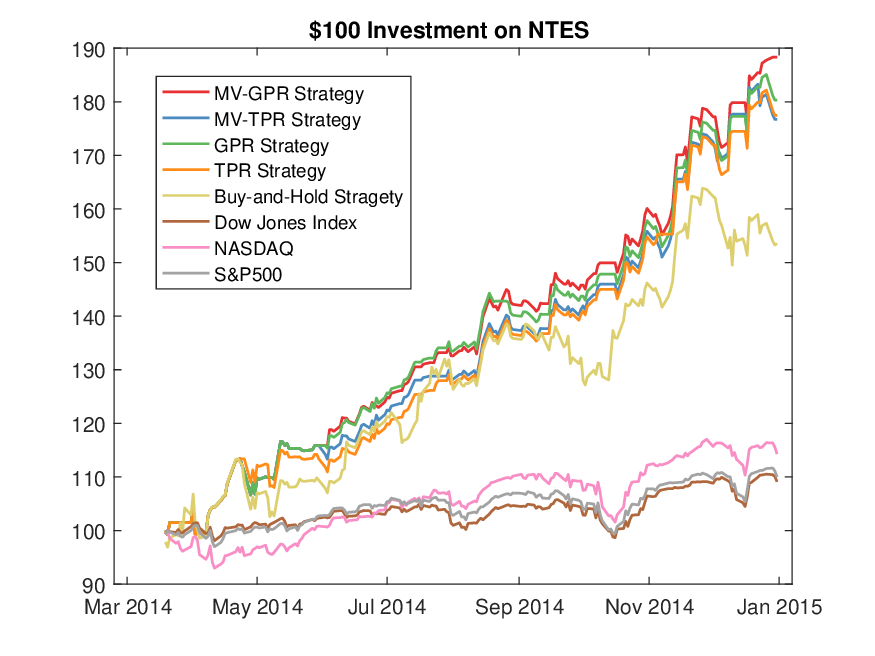}
\includegraphics[width=0.48\textwidth]{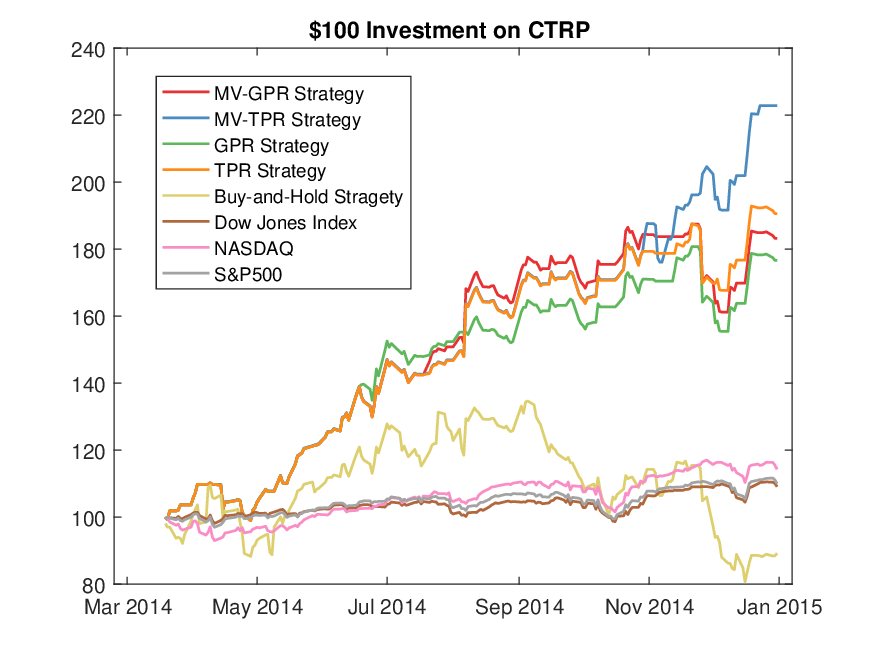}
\caption{The movement of invested \$100 in 200 days for 3 Chinese stocks in the US market. The top 4 lines in legend are Buy\&Sell strategies based on 4 prediction models, MV-GPR, MV-TPR, GPR, and TPR, respectively. The last 4 lines are Buy\&Hold strategies for the stock and for the three indices, INDU, NASDAQ, and NDX, respectively}
\label{fig:ChinaStocks}
\end{figure}

We apply MV-GPR, MV-TPR, GPR and TPR strategies and the results are demonstrated in Figure \ref{fig:ChinaStocks}. Furthermore, Table \ref{tab:PeriodBIDU}, Table \ref{tab:PeriodCTRP} and Table \ref{tab:PeriodNTES} in the appendix summarize the results by period for each stock respectively. In particular, the Buy\&Sell signal examples  for each stock are shown in Table \ref{tab:StockBIDU}, Table \ref{tab:StockCTRP} and Table \ref{tab:StockNTES} respectively, along with other relevant details.

From the view of Figure \ref{fig:ChinaStocks}, there is no doubt that a \$100 investment for each stock has sharply increased over 200 days period using Buy\&Sell strategies no matter whether the stock went up or down during this period. In particular, the stock prices of BIDU and NTES rose up gradually while CTRP hit the peak and then decreased in a large scale. Anyway, the Buy\&Sell strategies based on different prediction models have still achieved considerable profits compared with Buy\&Hold strategies for the corresponding investments. However, the different prediction models have diverse performances for each stock.  For BIDU, GPR-based models, including MV-GPR and GPR, outperform TPR-based models, including MV-TPR and TPR. For NTES, all the models for Buy\&Sell strategy have a similar performance. Admittedly, TPR-based models, especially MV-TPR, have an outstanding performance for stock CTRP.

\subsubsection{Diverse sectors in Dow 30}\label{Section:Dow30}
Owing to the globalization of capital, there has been a significant shift in the relative importance of national and economic influences in the world's largest equity markets and the impact of industrial sector effects is now gradually replacing that of country effects in these markets \cite{baca2000rise}. Therefore, a further example is carried out under the diverse industrial sectors in Dow 30 from New York Stock Exchange (NYSE) and NASDAQ.

Initially, the classification of stocks based on diverse industrial sectors in Dow 30 has to be done. There are two main industry classification taxonomies, including Industry Classification Benchmark (ICB) and Global Industry Classification Standard (GICS). In our research, ICB is used to segregate markets into sectors within the macroeconomy. The stocks in Dow 30 are classified in Table \ref{tab:sectorclassification}.
\begin{table}[htbp]
  \scriptsize
  \setlength{\abovecaptionskip}{0pt}
  \setlength{\belowcaptionskip}{5pt}
  \centering
  \caption{Stock components of Dow 30}
    \begin{tabular}{clcll}
    \toprule
    \textbf{Ticker} & \textbf{Company} & \textbf{Exchange} & \textbf{Industry} & \textbf{Industry} \footnotemark (ICB) \\
    \midrule
    DD    & DuPont & NYSE  & Chemical industry & Basic Materials \\
    KO    & Coca-Cola & NYSE  & Beverages & Consumer Goods \\
    PG    & Procter \& Gamble & NYSE  & Consumer goods & Consumer Goods \\
    MCD   & McDonald's & NYSE  & Fast food & Consumer Goods \\
    NKE   & Nike  & NYSE  & Apparel & Consumer Services \\
    DIS   & Walt Disney & NYSE  & Broadcasting and entertainment & Consumer Services \\
    HD    & The Home Depot & NYSE  & Home improvement retailer & Consumer Services \\
    WMT   & Wal-Mart & NYSE  & Retail & Consumer Services \\
    JPM   & JPMorgan Chase & NYSE  & Banking & Financials \\
    GS    & Goldman Sachs & NYSE  & Banking, Financial services & Financials \\
    V     & Visa  & NYSE  & Consumer banking & Financials \\
    AXP   & American Express & NYSE  & Consumer finance & Financials \\
    TRV   & Travelers & NYSE  & Insurance & Financials \\
    UNH   & UnitedHealth Group & NYSE  & Managed health care & Health Care \\
    JNJ   & Johnson \& Johnson & NYSE  & Pharmaceuticals & Health Care \\
    MRK   & Merck & NYSE  & Pharmaceuticals & Health Care \\
    PFE   & Pfizer & NYSE  & Pharmaceuticals & Health Care \\
    BA    & Boeing & NYSE  & Aerospace and defense & Industrials \\
    MMM   & 3M    & NYSE  & Conglomerate & Industrials \\
    GE    & General Electric & NYSE  & Conglomerate & Industrials \\
    UTX   & United Technologies & NYSE  & Conglomerate & Industrials \\
    CAT   & Caterpillar & NYSE  & Construction and mining equipment & Industrials \\
    CVX   & Chevron & NYSE  & Oil \& gas & Oil \& Gas \\
    XOM   & ExxonMobil & NYSE  & Oil \& gas & Oil \& Gas \\
    CSCO  & Cisco Systems & NASDAQ & Computer networking & Technology \\
    IBM   & IBM   & NYSE  & Computers and technology & Technology \\
    AAPL  & Apple & NASDAQ & Consumer electronics & Technology \\
    INTC  & Intel & NASDAQ & Semiconductors & Technology \\
    MSFT  & Microsoft & NASDAQ & Software & Technology \\
    VZ    & Verizon & NYSE  & Telecommunication & Telecommunications \\
    \bottomrule
    \end{tabular}%
  \label{tab:sectorclassification}%
\end{table}%
\footnotetext{Note that the terms "industry" and "sector" are reversed from the Global Industry Classification Standard (GICS) taxonomy.} Due to the multivariate process models considering at least two related stocks in one group, the first (Basic Materials) and the last industrial sector (Telecommunications), each consisting of only one stock, are excluded. As a result, our experiments are performed seven times for the seven grouped industrial sector stocks, including Oil\&Gas, Industrial, Consumer Goods, Health Care, Consumer Services, Financials and Technology, respectively.

Secondly, the 4 models, MV-GPR, MV-TPR, GPR and TPR, are applied in the same way as in Section \ref{Section:Chinese} and the ranking of stock investment performance is listed in Table \ref{tab:SectorStock} (the details are summarized in Table \ref{tab:SectorStockDetail}).
On the whole, for each stock, there is no doubt that using Buy\&Sell strategy is much better than using Buy\&Hold strategy regardless of the industrial sector. Specifically, MV-GPR makes a satisfactory performance overall in the sectors,  Industrials, Consumer Services and Financials while MV-TPR  has a higher ranking in Health Care in general.
\begin{table}[htbp]
  \scriptsize
  \setlength{\abovecaptionskip}{0pt}
  \setlength{\belowcaptionskip}{5pt}
  \centering
  \caption{Stock investment performance ranking under different strategies}
    \begin{tabular}{c|c|c|c|c|c|cccc}
    \toprule
    \multicolumn{1}{c|}{\multirow{2}[1]{*}{\textbf{Ticker}}} & \multicolumn{1}{c|}{\multirow{2}[1]{*}{\textbf{Industry}}} & \multicolumn{4}{c|}{\textbf{Buy\&Sell Strategy}} & \multicolumn{4}{c}{\textbf{Buy\&Hold Stragegy}} \\
    \cmidrule(r){3-10}
    \multicolumn{1}{c|}{} & \multicolumn{1}{c|}{} & \textbf{MV-GPR} & \textbf{MV-TPR} & \textbf{GPR}   & \textbf{TPR}   & \textbf{Stock} & \textbf{INDU}  & \textbf{NDX}   & \textbf{SPX} \\
    \midrule
    CVX   & Oil \& Gas & 3rd   & 4th   & 2nd   & \textit{\textbf{1st}} & 8th   & 7th   & 5th   & 6th \\
    XOM   & Oil \& Gas & 4th   & 2nd   & 3rd   & \textit{\textbf{1st}} & 8th   & 7th   & 5th   & 6th \\
        \midrule
    MMM   & Industrials & 2nd   & 3rd   & \textit{\textbf{1st}} & 4th   & 5th   & 8th   & 6th   & 7th \\
    BA    & Industrials & \textit{\textbf{1st}} & 2nd   & 3rd   & 4th   & 8th   & 7th   & 5th   & 6th \\
    CAT   & Industrials & 3rd   & 4th   & 2nd   & \textit{\textbf{1st}} & 8th   & 7th   & 5th   & 6th \\
    GE    & Industrials & 2nd   & 4th   & 3rd   & \textit{\textbf{1st}} & 8th   & 7th   & 5th   & 6th \\
    UTX   & Industrials & 2nd   & 4th   & 3rd   & \textit{\textbf{1st}} & 8th   & 7th   & 5th   & 6th \\
        \midrule
    KO    & Consumer Goods & 2nd   & \textit{\textbf{1st}} & 3rd   & 4th   & 6th   & 8th   & 5th   & 7th \\
    MCD   & Consumer Goods & 2nd   & 4th   & \textit{\textbf{1st}} & 3rd   & 8th   & 7th   & 5th   & 6th \\
    PG    & Consumer Goods & 3rd   & 4th   & \textit{\textbf{1st}} & 2nd   & 5th   & 8th   & 6th   & 7th \\
        \midrule
    JNJ   & Health Care & 3rd   & 2nd   & \textit{\textbf{1st}} & 4th   & 6th   & 8th   & 5th   & 7th \\
    MRK   & Health Care & 3rd   & 2nd   & 4th   & \textit{\textbf{1st}} & 8th   & 7th   & 5th   & 6th \\
    PFE   & Health Care & 4th   & \textit{\textbf{1st}} & 3rd   & 2nd   & 8th   & 7th   & 5th   & 6th \\
    UNH   & Health Care & 2nd   & 3rd   & \textit{\textbf{1st}} & 4th   & 5th   & 8th   & 6th   & 7th \\
       \midrule
    HD    & Consumer Services & \textit{\textbf{1st}} & 4th   & 3rd   & 2nd   & 5th   & 8th   & 6th   & 7th \\
    NKE   & Consumer Services & 2nd   & 3rd   & 4th   & \textit{\textbf{1st}} & 5th   & 8th   & 6th   & 7th \\
    WMT   & Consumer Services & \textit{\textbf{1st}} & 4th   & 3rd   & 2nd   & 5th   & 8th   & 6th   & 7th \\
    DIS   & Consumer Services & 3rd   & 2nd   & \textit{\textbf{1st}} & 4th   & 5th   & 8th   & 6th   & 7th \\
        \midrule
    AXP   & Financials & 2nd   & 4th   & \textit{\textbf{1st}} & 3rd   & 8th   & 7th   & 5th   & 6th \\
    GS    & Financials & 2nd   & \textit{\textbf{1st}} & 3rd   & 4th   & 5th   & 8th   & 6th   & 7th \\
    JPM   & Financials & 2nd   & 4th   & \textit{\textbf{1st}} & 3rd   & 6th   & 8th   & 5th   & 7th \\
    TRV   & Financials & 2nd   & 3rd   & \textit{\textbf{1st}} & 4th   & 5th   & 8th   & 6th   & 7th \\
    V     & Financials & \textit{\textbf{1st}} & 4th   & 3rd   & 2nd   & 5th   & 8th   & 6th   & 7th \\
        \midrule
    AAPL  & Technology & 4th   & 2nd   & 3rd   & \textit{\textbf{1st}} & 5th   & 8th   & 6th   & 7th \\
    CSCO  & Technology & 2nd   & \textit{\textbf{1st}} & 3rd   & 4th   & 5th   & 8th   & 6th   & 7th \\
    IBM   & Technology & 4th   & \textit{\textbf{1st}} & 2nd   & 3rd   & 8th   & 7th   & 5th   & 6th \\
    INTC  & Technology & 3rd   & 4th   & 2nd   & \textit{\textbf{1st}} & 5th   & 8th   & 6th   & 7th \\
    MSFT  & Technology & 2nd   & 4th   & \textit{\textbf{1st}} & 3rd   & 5th   & 8th   & 6th   & 7th \\
    \bottomrule
    \end{tabular}%
  \label{tab:SectorStock}%
\end{table}%

Further analysis is considered using industrial sector portfolios, which consists of these grouped stocks by the same weight investment on each stock. For example, the Oil \& Gas portfolio investment is \$100 with \$50 shares CVX and \$50 shares XOM while the Technology portfolio investment is \$100 with the same \$20 investment on each stock in the industrial sector Technology. The diverse industry portfolio investment performance ranking is listed in Table \ref{tab:SectorPortfolio} (the details are described in Table \ref{tab:SectorPortfolioDetail}).
Apparently, the Buy\&Sell strategies performed better than the Buy\&Hold strategies. MV-GPR suits better in three industries, including Consumer Goods, Consumer Services, and Financials, followed by TPR which performed best in Oil\&Gas and Industrials. The optimal investment strategy in Health Care is MV-TPR while in Technology industry, using GPR seems to be the most profitable.
\begin{table}[htbp]
  \scriptsize
  \setlength{\abovecaptionskip}{0pt}
  \setlength{\belowcaptionskip}{5pt}
  \centering
  \caption{Industry portfolio investment performance ranking under different strategies }
    \begin{tabular}{c|cccc|cccc}
    \toprule
    \multicolumn{1}{c|}{\multirow{2}[1]{*}{\textbf{Industry Portfolio}}} & \multicolumn{4}{c|}{\textbf{Buy\&Sell Strategy}} & \multicolumn{4}{c}{\textbf{Buy\&Hold Stragegy}} \\
    \cmidrule(r){2-9}
    \multicolumn{1}{c|}{} & \textbf{MV-GPR} & \textbf{MV-TPR} & \textbf{GPR}   & \textbf{TPR}   & \textbf{Stock} & \textbf{INDU}  & \textbf{NDX}   & \textbf{SPX} \\
    \midrule
    Oil \& Gas & 4th   & 3rd   & 2nd   & \textit{\textbf{1st}} & 8th   & 7th   & 5th   & 6th \\
        \midrule
    Industrials & 2nd   & 4th   & 3rd   & \textit{\textbf{1st}} & 8th   & 7th   & 5th   & 6th \\
        \midrule
    Consumer Goods & \textit{\textbf{1st}} & 4th   & 2nd   & 3rd   & 7th   & 8th   & 5th   & 6th \\
        \midrule
    Health Care & 4th   & \textit{\textbf{1st}} & 3rd   & 2nd   & 6th   & 8th   & 5th   & 7th \\
       \midrule
    Consumer Services & \textit{\textbf{1st}} & 4th   & 3rd   & 2nd   & 5th   & 8th   & 6th   & 7th \\
        \midrule
    Financials & \textit{\textbf{1st}} & 4th   & 2nd   & 3rd   & 5th   & 8th   & 6th   & 7th \\
        \midrule
    Technology & 4th   & 3rd   & \textit{\textbf{1st}} & 2nd   & 5th   & 8th   & 6th   & 7th \\
    \bottomrule
    \end{tabular}%
  \label{tab:SectorPortfolio}%
\end{table}%
\section{Conclusion and discussion}\label{sec:conclusion}
In this paper we have proposed a unified framework for multi-output regression and
prediction. Using this framework, we introduced a novel multivariate Student$-t$ process regression model (MV-TPR) and also reformulated the multivariate Gaussian process
regression (MV-GPR) which overcomes some limitations of the existing methods.
It could also be used to derive regression models of
general elliptical processes. Under this framework,
the model settings, derivations and computations for both MV-GPR and MV-TPR
are all directly performed in matrix form. MV-GPR is a more straightforward method
compared to the existing vectorization method and
can be implemented in the same way as the conventional GPR. Similar to the existing Gaussian process regression for vector-valued function,
 our models are also able to learn the correlations between inputs and outputs,
but with more convenient and flexible formulations.
The proposed MV-TPR also possesses closed-form expressions for the marginal likelihood
and the predictive distributions under this unified framework. Thus
the same optimization approaches as used in the conventional GPR can be adopted.
The usefulness of the proposed methods is illustrated through several
numerical examples. It is empirically demonstrated that MV-TPR has superiority in
prediction in these examples, including the simulated examples, air quality prediction
and bike rent prediction.

The proposed methods are also applied to stock market modelling and are shown to have
the ability to make a profitable stock investment. For the three "Chinese concept stocks",
the Buy\&Sell strategies based on the proposed models have more satisfactory performances, compared with the Buy\&Hold strategies for the corresponding stocks and
three main indices in the US; in particular, the strategy based on MV-TPR has outstanding returns for NetEase among three stocks. When applied to the industrial sectors in Dow 30,
the results indicate that the strategies based on MV-GPR have generally considerable performances in Industrials, Consumer Goods, Consumer Services, and Financials sectors,
 while those based on MV-TPR can make maximum profit in Health Care sector.

It is noted that we used the squared exponential kernel for demonstration
in all our experiments. However, it can be expected that
other kernels or more complicated kernels may lead to better results,
especially in the financial data examples, as the SE kernel may oversmooth data;
see \cite{cite:Rbook} for more details on the choice of kernels.
In this paper we assume that different outputs are observed at the same covariate values. In practice, different responses may be observed at different locations.
These cases are, however, difficult for the proposed framework since all the outputs have to be considered as a matrix in our models, rather than as a vector with adjustable length.
Another issue worth noting is that, as discussed in Section 4.3.2,
in our proposed framework all the outputs are considered as a whole and the same kernel
(and hyperparameters) is used for all the outputs, which may not be appropriate if different outputs have very different patterns such as in the air quality example.
Moreover, it is also important to further study how to improve the quality
of the parameter estimates, for example
using optimal design methods for parameter estimation as discussed in \cite{boukouvalas2014optimal}.
All of these problems are worth further investigation and exploration and will be our future works.

\bibliographystyle{unsrtnat}
\bibliography{reference.bib}

\begin{thebibliography}{30}
\providecommand{\natexlab}[1]{#1}
\providecommand{\url}[1]{\texttt{#1}}
\expandafter\ifx\csname urlstyle\endcsname\relax
  \providecommand{\doi}[1]{doi: #1}\else
  \providecommand{\doi}{doi: \begingroup \urlstyle{rm}\Url}\fi

\bibitem[Rasmussen(1999)]{cite:RCE}
Carl~Edward Rasmussen.
\newblock \emph{Evaluation of {G}aussian processes and other methods for
  non-linear regression}.
\newblock University of Toronto, 1999.

\bibitem[Boyle and Frean(2005)]{boyle2005dependent}
Phillip Boyle and Marcus Frean.
\newblock Dependent gaussian processes.
\newblock In \emph{Advances in neural information processing systems}, pages
  217--224, 2005.

\bibitem[Neal(2012)]{cite:neal}
Radford~M Neal.
\newblock \emph{Bayesian learning for neural networks}, volume 118.
\newblock Springer Science \& Business Media, 2012.

\bibitem[Williams(1997)]{williams1997computing}
Christopher~KI Williams.
\newblock Computing with infinite networks.
\newblock \emph{Advances in neural information processing systems}, pages
  295--301, 1997.

\bibitem[MacKay(1997)]{cite:MacKay}
David~JC MacKay.
\newblock Gaussian processes-a replacement for supervised neural networks?
\newblock 1997.

\bibitem[Williams and Rasmussen(1996)]{cite:WCKI}
Christopher~KI Williams and Carl~Edward Rasmussen.
\newblock Gaussian processes for regression.
\newblock In \emph{Advances in neural information processing systems}, pages
  514--520, 1996.

\bibitem[Brahim-Belhouari and Vesin(2001)]{cite:BB}
Sofiane Brahim-Belhouari and Jean-Marc Vesin.
\newblock Bayesian learning using {G}aussian process for time series
  prediction.
\newblock In \emph{Statistical Signal Processing, 2001. Proceedings of the 11th
  IEEE Signal Processing Workshop on}, pages 433--436. IEEE, 2001.

\bibitem[Brahim-Belhouari and Bermak(2004)]{cite:BBS}
Sofiane Brahim-Belhouari and Amine Bermak.
\newblock Gaussian process for nonstationary time series prediction.
\newblock \emph{Computational Statistics and Data Analysis}, 47\penalty0
  (4):\penalty0 705--712, 2004.

\bibitem[Wang and Chen(2015)]{wang2015gaussian}
Bo~Wang and Tao Chen.
\newblock Gaussian process regression with multiple response variables.
\newblock \emph{Chemometrics and Intelligent Laboratory Systems}, 142:\penalty0
  159--165, 2015.

\bibitem[Chakrabarty et~al.(2015)Chakrabarty, Biswas, Bhattacharya,
  et~al.]{chakrabarty2015bayesian}
Dalia Chakrabarty, Munmun Biswas, Sourabh Bhattacharya, et~al.
\newblock Bayesian nonparametric estimation of {M}ilky {W}ay parameters using
  matrix-variate data, in a new gaussian process based method.
\newblock \emph{Electronic Journal of Statistics}, 9\penalty0 (1):\penalty0
  1378--1403, 2015.

\bibitem[Alvarez et~al.(2012)Alvarez, Rosasco, Lawrence,
  et~al.]{alvarez2011kernels}
Mauricio~A Alvarez, Lorenzo Rosasco, Neil~D Lawrence, et~al.
\newblock Kernels for vector-valued functions: {A} review.
\newblock \emph{Foundations and Trends{\textregistered} in Machine Learning},
  4\penalty0 (3):\penalty0 195--266, 2012.

\bibitem[Gupta and Nagar(1999)]{gupta1999matrix}
Arjun~K Gupta and Daya~K Nagar.
\newblock \emph{Matrix variate distributions}, volume 104.
\newblock CRC Press, 1999.

\bibitem[Dawid(1981)]{dawid1981some}
A~Philip Dawid.
\newblock Some matrix-variate distribution theory: notational considerations
  and a bayesian application.
\newblock \emph{Biometrika}, 68\penalty0 (1):\penalty0 265--274, 1981.

\bibitem[Zhu et~al.(2008)Zhu, Yu, and Gong]{zhu2008predictive}
Shenghuo Zhu, Kai Yu, and Yihong Gong.
\newblock Predictive matrix-variate t models.
\newblock In \emph{Advances in Neural Information Processing Systems}, pages
  1721--1728, 2008.

\bibitem[Zhang(2006)]{zhang2006schur}
Fuzhen Zhang.
\newblock \emph{The {S}chur complement and its applications}, volume~4.
\newblock Springer Science \& Business Media, 2006.

\bibitem[Shah et~al.(2014)Shah, Wilson, and Ghahramani]{shah2014student}
Amar Shah, Andrew~Gordon Wilson, and Zoubin Ghahramani.
\newblock Student-t processes as alternatives to {G}aussian processes.
\newblock In \emph{AISTATS}, pages 877--885, 2014.

\bibitem[Rasmussen and Williams(2006)]{cite:Rbook}
Carl~Edward Rasmussen and Christopher~KI Williams.
\newblock \emph{Gaussian processes for machine learning}, volume~1.
\newblock MIT press Cambridge, 2006.

\bibitem[Roberts et~al.(2013)Roberts, Osborne, Ebden, Reece, Gibson, and
  Aigrain]{cite:twins}
Stephen Roberts, Michael Osborne, Mark Ebden, Steven Reece, Neale Gibson, and
  Suzanne Aigrain.
\newblock Gaussian processes for time-series modelling.
\newblock \emph{Phil. Trans. R. Soc. A}, 371\penalty0 (1984):\penalty0
  20110550, 2013.

\bibitem[Neal(1997)]{cite:monte}
Radford~M Neal.
\newblock Monte carlo implementation of {G}aussian process models for bayesian
  regression and classification.
\newblock \emph{arXiv preprint physics/9701026}, 1997.

\bibitem[MacKay(1998)]{mackay1998introduction}
David~JC MacKay.
\newblock Introduction to {G}aussian processes.
\newblock \emph{NATO ASI Series F Computer and Systems Sciences}, 168:\penalty0
  133--166, 1998.

\bibitem[Duvenaud et~al.(2013)Duvenaud, Lloyd, Grosse, Tenenbaum, and
  Ghahramani]{cite:autokernel}
David Duvenaud, James~Robert Lloyd, Roger Grosse, Joshua~B Tenenbaum, and
  Zoubin Ghahramani.
\newblock Structure discovery in nonparametric regression through compositional
  kernel search.
\newblock \emph{arXiv preprint arXiv:1302.4922}, 2013.

\bibitem[Wilson(2014)]{wilson2014covariance}
Andrew~Gordon Wilson.
\newblock \emph{Covariance kernels for fast automatic pattern discovery and
  extrapolation with {G}aussian processes}.
\newblock PhD thesis, PhD thesis, University of Cambridge, 2014.

\bibitem[Chen and Wang(2018)]{chen2018priors}
Zexun Chen and Bo~Wang.
\newblock How priors of initial hyperparameters affect gaussian process
  regression models.
\newblock \emph{Neurocomputing}, 275:\penalty0 1702--1710, 2018.

\bibitem[Boukouvalas et~al.(2014)Boukouvalas, Cornford, and
  Stehl{\'\i}k]{boukouvalas2014optimal}
Alexis Boukouvalas, Dan Cornford, and Milan Stehl{\'\i}k.
\newblock Optimal design for correlated processes with input-dependent noise.
\newblock \emph{Computational Statistics \& Data Analysis}, 71:\penalty0
  1088--1102, 2014.

\bibitem[Fanaee-T and Gama(2014)]{fanaee2014event}
Hadi Fanaee-T and Joao Gama.
\newblock Event labeling combining ensemble detectors and background knowledge.
\newblock \emph{Progress in AI}, 2\penalty0 (2-3):\penalty0 113--127, 2014.

\bibitem[De~Vito et~al.(2008)De~Vito, Massera, Piga, Martinotto, and
  Di~Francia]{de2008field}
S~De~Vito, E~Massera, M~Piga, L~Martinotto, and G~Di~Francia.
\newblock On field calibration of an electronic nose for benzene estimation in
  an urban pollution monitoring scenario.
\newblock \emph{Sensors and Actuators B: Chemical}, 129\penalty0 (2):\penalty0
  750--757, 2008.

\bibitem[Akbilgic et~al.(2014)Akbilgic, Bozdogan, and
  Balaban]{akbilgic2014novel}
Oguz Akbilgic, Hamparsum Bozdogan, and M~Erdal Balaban.
\newblock A novel hybrid {RBF} neural networks model as a forecaster.
\newblock \emph{Statistics and Computing}, 24\penalty0 (3):\penalty0 365--375,
  2014.

\bibitem[Luo et~al.(2012)Luo, Fang, and Esqueda]{luo2012overseas}
Yongli Luo, Fang Fang, and Omar~A Esqueda.
\newblock The overseas listing puzzle: {Post-IPO} performance of {C}hinese
  stocks and adrs in the {US} market.
\newblock \emph{Journal of multinational financial management}, 22\penalty0
  (5):\penalty0 193--211, 2012.

\bibitem[Baca et~al.(2000)Baca, Garbe, and Weiss]{baca2000rise}
Sean~P Baca, Brian~L Garbe, and Richard~A Weiss.
\newblock The rise of sector effects in major equity markets.
\newblock \emph{Financial Analysts Journal}, 56\penalty0 (5):\penalty0 34--40,
  2000.

\bibitem[Gentle(2007)]{gentle2007matrix}
James~E Gentle.
\newblock \emph{Matrix algebra: theory, computations, and applications in
  statistics}.
\newblock Springer Science \& Business Media, 2007.

\end{thebibliography}

\appendix
\section{Negative Log Marginal Likelihood and Gradient Evaluation}
\subsection{Matrix derivatives}
According to the chain rule of derivatives of matrix, there exists \cite{gentle2007matrix}:
Letting $U = f(X)$, the derivatives of the function $g(U)$ with respect to X are
$$
\frac{\partial g(U)}{\partial X_{ij}} = \tr\left[\left(\frac{\partial g(U)}{\partial U}\right)^{\mathrm{T}} \frac{\partial U}{\partial X_{ij}}\right],
$$
where X is an $n \times m$ matrix. Additionally, there are another two useful formulas of derivative with respect to X.
$$
\frac{\partial \ln \det(X)}{\partial X} = X^{-\mathrm{T}}, \quad \frac{\partial}{\partial X}\tr(AX^{-1}B) = -(X^{-1}BAX^{-1})^{\mathrm{T}},
$$
where X is an $n \times n$ matrix, A is a constant $m \times n$ matrix and B is a constant $n \times m$ matrix.
\subsection{Multivariate Gaussian process regression}\label{app:MV-GPR-Gradient}
For a matrix-variate observations $Y \sim \mathcal{MN}_{n,d}(M,\Sigma,\Omega)$ where $M \in \mathds{R}^{n \times d}, \Sigma \in \mathds{R}^{n \times n}, \Omega \in \mathds{R}^{d \times d}$, the negative log likelihood is
\begin{equation}
  \mathcal{L} = \frac{nd}{2}\ln(2\pi) + \frac{d}{2}\ln \det(\Sigma) + \frac{n}{2}\ln \det(\Omega) + \frac{1}{2}\tr(\Sigma^{-1}(Y-M)\Omega^{-1}(Y-M)^{\mathrm{T}}),
\end{equation}
where actually $\Sigma = K + \sigma_n^2 \mathbf{I}$ As we know there are several parameters in the kernel $k$ so that we can denote $K = K_{\theta}$. The parameter set denotes $\Theta = \{\theta_1, \theta_2, \ldots\}$. Besides, we denote the parameter matrix $\Omega = \Phi \Phi^{\mathrm{T}}$ since $\Omega$ is positive semi-definite, where
$$
\Phi =
\left[
\begin{matrix}
 \phi_{11}      & 0      & \cdots & 0      \\
 \phi_{21}      & \phi_{22}      & \cdots & 0      \\
 \vdots & \vdots & \ddots & \vdots \\
 \phi_{d1}      & \phi_{d2}      & \cdots & \phi_{dd}      \\
\end{matrix}
\right].
$$
To guarantee the uniqueness of $\Phi$, the diagonal elements are restricted to be positive and denote $\varphi_{ii} = \ln(\phi_{ii})$ for $i = 1,2,\cdots,d$. Therefore,
$$
\frac{\partial \Sigma}{\partial \sigma_n^2} = \mathbf{I}_n,  \quad \frac{\partial \Sigma
}{\partial \theta_i} = \frac{\partial K'_{\theta}}{\partial \theta_i},\qquad \frac{\partial \Omega}{\partial \phi_{ij}} = \mathbf{E}_{ij}\Phi^{\mathrm{T}} + \Phi \mathbf{E}_{ij}, \quad \frac{\partial \Omega}{\partial \varphi_{ii}} = \mathbf{J}_{ii}\Phi^{\mathrm{T}} + \Phi \mathbf{J}_{ii},
$$
where $\mathbf{E}_{ij}$ is the $d \times d$ elementary matrix having unity in the (i,j)-th element and zeros elsewhere, and $\mathbf{J}_{ii}$ is the same as $\mathbf{E}_{ij}$ but with the unity being replaced by $e^{\varphi_{ii}}$.

The derivatives of the negative log likelihood with respect to $\sigma_n^2$, $\theta_i$, $\phi_{ij}$ and $\varphi_{ii}$ are as follows.
The derivative with respect to $\theta_i$ is
\begin{eqnarray}\label{deriativeTheta}
  \frac{\partial \mathcal{L}}{\partial \theta_i} &=& \frac{d}{2}\frac{\partial \ln \det(\Sigma)}{\partial \theta_i} + \frac{1}{2}\frac{\partial}{\partial \theta_i}\tr(\Sigma^{-1}(Y-M)\Omega^{-1}(Y-M)^{\mathrm{T}}) \nonumber \\
   &=& \frac{d}{2} \tr\left[\left(\frac{\partial \ln \det(\Sigma)}{\partial \Sigma}\right)^{\mathrm{T}} \frac{\partial \Sigma}{\partial \theta_i}\right] +  \frac{1}{2} \tr\left[\left(\frac{\partial \tr(\Sigma^{-1}G)}{\partial \Sigma}\right)^{\mathrm{T}} \frac{\partial \Sigma}{\partial \theta_i}\right]  \nonumber \\
   &=& \frac{d}{2}\tr\left( \Sigma^{-1}\frac{\partial K'_{\theta}}{\partial \theta_i} \right) - \frac{1}{2}\tr\left(\Sigma^{-1}G\Sigma^{-1} \frac{\partial K'_{\theta}}{\partial \theta_i} \right) \nonumber \\
   &=& \frac{d}{2}\tr\left(\Sigma^{-1}\frac{\partial K'_{\theta}}{\partial \theta_i}\right) - \frac{1}{2}\tr\left(\alpha_{\Sigma}\Omega^{-1}\alpha_{\Sigma}^{\mathrm{T}}\frac{\partial K'_{\theta}}{\partial \theta_i}\right),
\end{eqnarray}
where $G =(Y-M)\Omega^{-1}(Y-M)^{\mathrm{T}} $ and $\alpha_{\Sigma} = \Sigma^{-1}(Y-M)$.The fourth equality is due to the symmetry of $\Sigma$.

Due to $\partial \Sigma / \partial \sigma^2_n = \mathbf{I}_n$, the derivative with respect to $\sigma_n^2$ is:
\begin{eqnarray}\label{deriativesigma2}
  \frac{\partial \mathcal{L}}{\partial \sigma_n^2}
   &=& \frac{d}{2}\tr(\Sigma^{-1}) - \frac{1}{2}\tr(\alpha_{\Sigma}\Omega^{-1}\alpha_{\Sigma}^{\mathrm{T}}).
\end{eqnarray}
Letting $\alpha_{\Omega} = \Omega^{-1}(Y-M)^{\mathrm{T}}$, the derivative with respect to $\phi_{ij}$ is
\begin{eqnarray}\label{deriativePhi}
  \frac{\partial \mathcal{L}}{\partial \phi_{ij}} &=& \frac{n}{2}\frac{\partial \ln \det(\Omega)}{\partial \phi_{ij}} + \frac{1}{2}\frac{\partial }{\partial \phi_{ij}}\tr(\Sigma^{-1}(Y-M)\Omega^{-1}(Y-M)^{\mathrm{T}}) \nonumber \\
   &=& \frac{n}{2} \tr\left(\Omega^{-1} \frac{\partial \Omega}{\partial \phi_{ij}}\right) -  \frac{1}{2} \tr\left[( (\Omega^{-1}(Y-M)^{\mathrm{T}}\Sigma^{-1}(Y-M)\Omega^{-1})^{\mathrm{T}})^{\mathrm{T}} \frac{\partial \Omega}{\partial \phi_{ij}}\right]  \nonumber \\
   &=& \frac{n}{2}\tr\left(\Omega^{-1}\frac{\partial \Omega}{\partial \phi_{ij}}\right) - \frac{1}{2}\tr\left(\alpha_{\Omega}\Sigma^{-1}\alpha_{\Omega}^{\mathrm{T}}\frac{\partial \Omega}{\partial \phi_{ij}}\right) \nonumber \\
   &=& \frac{n}{2}\tr[\Omega^{-1}(\mathbf{E}_{ij}\Phi^{\mathrm{T}} + \Phi \mathbf{E}_{ij})] - \frac{1}{2}\tr[\alpha_{\Omega}\Sigma^{-1}\alpha_{\Omega}^{\mathrm{T}}(\mathbf{E}_{ij}\Phi^{\mathrm{T}} + \Phi \mathbf{E}_{ij})],
\end{eqnarray}
where the third equation is due to the symmetry of $\Omega$. Similarly, the derivative with respect to $\varphi_{ii}$ is
\begin{eqnarray}\label{deriativePsi}
  \frac{\partial \mathcal{L}}{\partial \varphi_{ii}} &=& \frac{n}{2}\frac{\partial \ln \det(\Omega)}{\partial \varphi_{ii}} + \frac{1}{2}\frac{\partial }{\partial \varphi_{ii}}\tr(\Sigma^{-1}(Y-M)\Omega^{-1}(Y-M)^{\mathrm{T}}) \nonumber \\
  &=& \frac{n}{2}\tr[\Omega^{-1}(\mathbf{J}_{ii}\Phi^{\mathrm{T}} + \Phi \mathbf{J}_{ii})] - \frac{1}{2}\tr[\alpha_{\Omega}\Sigma^{-1}\alpha_{\Omega}^{\mathrm{T}}(\mathbf{J}_{ii}\Phi^{\mathrm{T}} + \Phi \mathbf{J}_{ii})].
\end{eqnarray}
\subsection{Multivariate Student\texorpdfstring{$-t$}{t} process regression}\label{app:MV-TPR-Gradient}
The negative log likelihood of observations $Y \sim \mathcal{MT}_{n,d}(\nu, M, \Sigma, \Omega)$ where $M \in \mathds{R}^{n \times d}, \Sigma \in \mathds{R}^{n \times n}, \Omega \in \mathds{R}^{d \times d}$, is
\begin{eqnarray*}
  \mathcal{L}  &=& \frac{1}{2}(\nu+ d+n -1) \ln \det(\mathbf{I}_n + \Sigma^{-1}(Y-M)\Omega^{-1}(Y-M)^{\mathrm{T}})  \\
               & & + \frac{d}{2}\ln \det(\Sigma) + \frac{n}{2}\ln \det(\Omega) + \ln\Gamma_n \left(\frac{1}{2}(\nu + n -1)\right)  + \frac{1}{2}dn\ln\pi  \\
               & & - \ln \Gamma_n \left(\frac{1}{2}(\nu + d + n -1)\right)  \\
               &=& \frac{1}{2}(\nu+ d+n -1) \ln \det(\Sigma +(Y-M)\Omega^{-1}(Y-M)^{\mathrm{T}}) - \frac{\nu + n -1}{2}\ln \det(\Sigma)  \\
               & & + \ln\Gamma_n \left(\frac{1}{2}(\nu + n -1)\right) - \ln \Gamma_n \left(\frac{1}{2}(\nu + d + n -1)\right) \\
               & & + \frac{n}{2}\ln \det(\Omega)+ \frac{1}{2}dn\ln\pi .
\end{eqnarray*}

Letting $U = \Sigma + (Y-M)\Omega^{-1}(Y-M)^{\mathrm{T}}$ and $\alpha_{\Omega} = \Omega^{-1}(Y-M)^{\mathrm{T}}$, the derivative of U with respect to $\sigma_n^2$, $\theta_i$,$\nu$, $\phi_{ij}$ and $\varphi_{ii}$ are
\begin{equation}\label{deriative3}
  \frac{\partial U}{\partial \sigma_n^2} = \mathbf{I}_n,\quad \frac{\partial U}{\partial \theta_i} = \frac{\partial K'_{\theta}}{\partial \theta_i}, \quad \frac{\partial U}{\partial \nu} =\mathrm{0},
\end{equation}
\begin{eqnarray}
  \frac{\partial U}{\partial \phi_{ij}} &=& - (Y-M)\Omega^{-1}\frac{\partial \Omega}{ \partial \phi_{ij}}\Omega^{-1}(Y-M)^{\mathrm{T}} = - \alpha_{\Omega}^{\mathrm{T}}\frac{\partial \Omega}{ \partial \phi_{ij}}\alpha_{\Omega} , \\
  \frac{\partial U}{\partial \varphi_{ii}} &=& - (Y-M)\Omega^{-1}\frac{\partial \Omega}{ \partial \varphi_{ii}}\Omega^{-1}(Y-M)^{\mathrm{T}} = - \alpha_{\Omega}^{\mathrm{T}}\frac{\partial \Omega}{ \partial \varphi_{ii}}\alpha_{\Omega}.
\end{eqnarray}
\\
Therefore, the derivative of negative log marginal likelihood with respect to $\theta_i$ is
\begin{eqnarray}\label{TderiativesTheta}
  \frac{\partial \mathcal{L}}{\partial \theta_i}  &=& \frac{(\tau + d)}{2}\frac{\partial \ln\det(U)}{\partial \theta_i}  - \frac{\tau}{2} \frac{\partial \ln \det(\Sigma)}{\partial \theta_i} \nonumber \\
   &=& \frac{(\tau+d)}{2}\tr\left(U^{-1} \frac{\partial K'_{\theta}}{\partial \theta_i}\right) - \frac{\tau}{2}\tr\left(\Sigma^{-1} \frac{\partial K'_{\theta}}{\partial \theta_i}\right),
\end{eqnarray}
where the constant $\tau = \nu + n -1$.

The derivative with respect to $\sigma_n^2$ is
\begin{equation}\label{TderiativesSigma2}
  \frac{\partial \mathcal{L}}{\partial \sigma^2_n}
   = \frac{(\tau+d)}{2}\tr(U^{-1}) - \frac{\tau}{2}\tr(\Sigma^{-1}).
\end{equation}
The derivative with respect to $\nu$ is
\begin{eqnarray}\label{TderivativeNu}
   \frac{\partial \mathcal{L}}{\partial \nu} = \frac{1}{2}\ln \det(U) - \frac{1}{2}\ln \det(\Sigma)  + \frac{1}{2}\psi_n(\frac{1}{2}\tau) - \frac{1}{2}\psi_n[\frac{1}{2}(\tau + d)]
\end{eqnarray}
where $\psi_n(\cdot)$ is the derivative of the function $\ln \Gamma_n(\cdot)$ with respect to $\nu$.

The derivative of $\mathcal{L}$ with respect to $\phi_{ij}$ is
\begin{eqnarray}\label{TderivativePhi}
  \frac{\partial \mathcal{L}}{\partial \phi_{ij}}  &=& \frac{(\tau + d)}{2}\frac{\partial \ln\det(U)}{\partial \phi_{ij}}  + \frac{n}{2} \frac{\partial \ln \det(\Omega)}{\partial \phi_{ij}} \nonumber \\
   &=& - \frac{(\tau +d)}{2}\tr[U^{-1}\alpha_{\Omega}^{\mathrm{T}}(\mathbf{E}_{ij}\Phi^{\mathrm{T}} + \Phi \mathbf{E}_{ij})\alpha_{\Omega}] + \frac{n}{2}\tr[\Omega^{-1}(\mathbf{E}_{ij}\Phi^{\mathrm{T}} + \Phi \mathbf{E}_{ij})].
\end{eqnarray}
Similarly, the derivative with respect to $\varphi_{ii}$ is
\begin{equation}\label{TderivativePsi}
  \frac{\partial \mathcal{L}}{\partial \varphi_{ii}}= -\frac{(\tau +d)}{2}\tr[U^{-1}\alpha_{\Omega}^{\mathrm{T}}(\mathbf{J}_{ii}\Phi^{\mathrm{T}} + \Phi \mathbf{J}_{ii})\alpha_{\Omega}] + \frac{n}{2}\tr[\Omega^{-1}(\mathbf{J}_{ii}\Phi^{\mathrm{T}} + \Phi \mathbf{J}_{ii})].
\end{equation}
\section{Three Chinese Stocks Investment Details}\label{Section:3ChineseStock}
\setcounter{table}{0}
\renewcommand{\thetable}{B\arabic{table}}

\begin{table}[htbp]
  \tiny
  \setlength{\abovecaptionskip}{0pt}
  \setlength{\belowcaptionskip}{5pt}
  \centering
  \caption{The movement of invested \$100 for 200 days split in to 20 periods (Stock: BIDU)}
    \begin{tabular}{c|c|c|c|c|cccc}
    \toprule
    \multicolumn{1}{c|}{\multirow{2}[0]{*}{\textbf{Forecast terms}}} & \multicolumn{4}{c|}{\textbf{Buy\&Sell decisions by prediction models}} & \multicolumn{4}{c}{\textbf{Buy\&Hold stock/index}} \\
    \cmidrule(r){2-9}
    \multicolumn{1}{c|}{} & \textbf{MV-GPR} & \textbf{MV-TPR} & \textbf{GPR}   & \textbf{TPR}   & \textbf{BIDU}  & \textbf{INDU}  & \textbf{NDX}   & \textbf{SPX} \\
    \midrule
    Beginning (\$) & \multicolumn{4}{c|}{100}       & \multicolumn{4}{c}{100} \\
    \midrule
    Period 1 & 103.53 & 100.97 & 103.53 & 100.97 & 97.14 & 101.20 & 98.70 & 100.71 \\
    Period 2 & 109.80 & 106.09 & 115.60 & 102.03 & 94.87 & 99.55 & 94.10 & 98.44 \\
    Period 3 & 108.37 & 104.71 & 115.96 & 100.71 & 93.89 & 101.50 & 96.64 & 100.62 \\
    Period 4 & 117.17 & 113.22 & 125.37 & 108.89 & 95.21 & 101.70 & 96.94 & 100.87 \\
    Period 5 & 127.84 & 123.52 & 136.79 & 118.80 & 102.22 & 102.22 & 100.79 & 102.55 \\
    Period 6 & 137.94 & 130.14 & 147.59 & 128.18 & 107.39 & 102.44 & 101.54 & 103.09 \\
    Period 7 & 146.20 & 137.93 & 156.43 & 135.86 & 112.11 & 103.12 & 103.25 & 104.54 \\
    Period 8 & 155.97 & 147.15 & 166.88 & 144.94 & 113.57 & 103.72 & 105.34 & 105.09 \\
    Period 9 & 177.94 & 167.88 & 175.27 & 152.21 & 138.22 & 103.82 & 106.98 & 105.67 \\
    Period 10 & 179.83 & 174.07 & 177.13 & 153.83 & 131.26 & 101.33 & 104.90 & 103.17 \\
    Period 11 & 179.08 & 173.35 & 176.05 & 153.19 & 130.71 & 104.07 & 109.34 & 106.20 \\
    Period 12 & 190.96 & 184.85 & 187.73 & 163.35 & 137.58 & 104.75 & 110.49 & 106.91 \\
    Period 13 & 201.08 & 194.63 & 204.44 & 177.89 & 131.12 & 105.12 & 109.57 & 106.52 \\
    Period 14 & 207.28 & 200.64 & 210.75 & 183.38 & 132.54 & 104.01 & 108.35 & 104.94 \\
    Period 15 & 210.54 & 203.80 & 214.06 & 186.26 & 132.19 & 100.39 & 104.41 & 101.70 \\
    Period 16 & 233.92 & 226.43 & 237.83 & 206.95 & 144.35 & 106.31 & 112.48 & 107.77 \\
    Period 17 & 252.47 & 233.71 & 256.69 & 223.36 & 148.99 & 108.03 & 113.68 & 109.03 \\
    Period 18 & 250.38 & 227.57 & 254.56 & 221.51 & 143.25 & 109.45 & 116.17 & 110.38 \\
    Period 19 & 260.24 & 223.38 & 264.59 & 230.23 & 134.23 & 104.49 & 110.33 & 105.37 \\
    Period 20 & 270.29 & 233.29 & 274.81 & 239.13 & 139.12 & 109.10 & 114.29 & 109.97 \\
    \bottomrule
    \end{tabular}%
  \label{tab:PeriodBIDU}%
\end{table}%
\begin{table}[htbp]
  \tiny
  \setlength{\abovecaptionskip}{0pt}
  \setlength{\belowcaptionskip}{5pt}
  \centering
  \caption{The movement of invested \$100 for 200 days split in to 20 periods (Stock: CTRP)}
    \begin{tabular}{c|c|c|c|c|cccc}
    \toprule
    \multicolumn{1}{c|}{\multirow{2}[0]{*}{\textbf{Forecast terms}}} & \multicolumn{4}{c|}{\textbf{Buy\&Sell decisions by prediction models}} & \multicolumn{4}{c}{\textbf{Buy\&Hold stock/index}} \\
    \cmidrule(r){2-9}
    \multicolumn{1}{c|}{} & \textbf{MV-GPR} & \textbf{MV-TPR} & \textbf{GPR}   & \textbf{TPR}   & \textbf{CTRP}  & \textbf{INDU}  & \textbf{NDX}   & \textbf{SPX} \\
    \midrule
    Beginning (\$) & \multicolumn{4}{c|}{100}       & \multicolumn{4}{c}{100} \\
    \midrule
    Period 1 & 105.67 & 105.67 & 105.67 & 105.67 & 100.78 & 101.20 & 98.70 & 100.71 \\
    Period 2 & 102.39 & 102.52 & 102.39 & 102.39 & 99.65 & 99.55 & 94.10 & 98.44 \\
    Period 3 & 102.77 & 102.90 & 102.77 & 102.77 & 91.63 & 101.50 & 96.64 & 100.62 \\
    Period 4 & 110.05 & 110.19 & 110.05 & 110.05 & 100.39 & 101.70 & 96.94 & 100.87 \\
    Period 5 & 121.51 & 121.66 & 121.51 & 121.51 & 108.33 & 102.22 & 100.79 & 102.55 \\
    Period 6 & 131.02 & 131.19 & 131.02 & 131.02 & 113.59 & 102.44 & 101.54 & 103.09 \\
    Period 7 & 138.90 & 139.08 & 144.25 & 138.90 & 120.90 & 103.12 & 103.25 & 104.54 \\
    Period 8 & 140.19 & 140.37 & 145.58 & 140.19 & 118.02 & 103.72 & 105.34 & 105.09 \\
    Period 9 & 150.29 & 146.38 & 151.82 & 146.20 & 131.35 & 103.82 & 106.98 & 105.67 \\
    Period 10 & 167.38 & 163.03 & 154.49 & 162.82 & 128.82 & 101.33 & 104.90 & 103.17 \\
    Period 11 & 166.33 & 162.01 & 154.28 & 161.80 & 127.37 & 104.07 & 109.34 & 106.20 \\
    Period 12 & 176.07 & 171.50 & 163.32 & 171.28 & 133.52 & 104.75 & 110.49 & 106.91 \\
    Period 13 & 176.00 & 171.43 & 163.25 & 171.21 & 117.53 & 105.12 & 109.57 & 106.52 \\
    Period 14 & 170.50 & 166.08 & 158.15 & 165.86 & 109.88 & 104.01 & 108.35 & 104.94 \\
    Period 15 & 178.68 & 174.04 & 165.73 & 173.82 & 108.35 & 100.39 & 104.41 & 101.70 \\
    Period 16 & 184.31 & 187.64 & 170.96 & 179.30 & 114.27 & 106.31 & 112.48 & 107.77 \\
    Period 17 & 183.77 & 191.85 & 176.72 & 180.87 & 115.96 & 108.03 & 113.68 & 109.03 \\
    Period 18 & 163.88 & 194.84 & 158.00 & 169.76 & 93.94 & 109.45 & 116.17 & 110.38 \\
    Period 19 & 169.89 & 201.99 & 163.80 & 176.73 & 80.69 & 104.49 & 110.33 & 105.37 \\
    Period 20 & 183.25 & 222.82 & 176.67 & 190.63 & 89.20 & 109.10 & 114.29 & 109.97 \\
    \bottomrule
    \end{tabular}%
  \label{tab:PeriodCTRP}%
\end{table}%
\begin{table}[htbp]
  \tiny
  \setlength{\abovecaptionskip}{0pt}
  \setlength{\belowcaptionskip}{5pt}
  \centering
  \caption{The movement of invested \$100 for 200 days split in to 20 periods (Stock: NTES)}
    \begin{tabular}{c|c|c|c|c|cccc}
    \toprule
    \multicolumn{1}{c|}{\multirow{2}[0]{*}{\textbf{Forecast terms}}} & \multicolumn{4}{c|}{\textbf{Buy\&Sell decisions by prediction models}} & \multicolumn{4}{c}{\textbf{Buy\&Hold stock/index}} \\
    \cmidrule(r){2-9}
    \multicolumn{1}{c|}{} & \textbf{MV-GPR} & \textbf{MV-TPR} & \textbf{GPR}   & \textbf{TPR}   & \textbf{NTES}  & \textbf{INDU}  & \textbf{NDX}   & \textbf{SPX} \\
    \midrule
    Beginning (\$) & \multicolumn{4}{c|}{100}       & \multicolumn{4}{c}{100} \\
    \midrule
    Period 1 & 104.51 & 104.51 & 104.51 & 104.51 & 106.79 & 101.20 & 98.70 & 100.71 \\
    Period 2 & 106.19 & 106.19 & 106.19 & 106.19 & 106.35 & 99.55 & 94.10 & 98.44 \\
    Period 3 & 106.80 & 106.80 & 106.80 & 109.12 & 104.14 & 101.50 & 96.64 & 100.62 \\
    Period 4 & 115.90 & 115.90 & 115.90 & 114.28 & 108.66 & 101.70 & 96.94 & 100.87 \\
    Period 5 & 115.82 & 115.82 & 115.82 & 114.21 & 109.84 & 102.22 & 100.79 & 102.55 \\
    Period 6 & 120.96 & 117.65 & 120.73 & 115.20 & 116.55 & 102.44 & 101.54 & 103.09 \\
    Period 7 & 123.27 & 121.54 & 124.72 & 119.01 & 120.32 & 103.12 & 103.25 & 104.54 \\
    Period 8 & 127.77 & 125.33 & 128.62 & 122.73 & 117.34 & 103.72 & 105.34 & 105.09 \\
    Period 9 & 133.21 & 128.81 & 134.09 & 127.95 & 128.53 & 103.82 & 106.98 & 105.67 \\
    Period 10 & 133.36 & 128.96 & 134.24 & 128.09 & 127.40 & 101.33 & 104.90 & 103.17 \\
    Period 11 & 141.13 & 136.47 & 142.80 & 135.56 & 134.83 & 104.07 & 109.34 & 106.20 \\
    Period 12 & 141.45 & 136.78 & 139.44 & 135.87 & 137.23 & 104.75 & 110.49 & 106.91 \\
    Period 13 & 145.98 & 141.16 & 143.90 & 140.22 & 134.27 & 105.12 & 109.57 & 106.52 \\
    Period 14 & 147.95 & 144.00 & 145.84 & 143.04 & 129.90 & 104.01 & 108.35 & 104.94 \\
    Period 15 & 151.75 & 147.70 & 149.59 & 146.71 & 139.19 & 100.39 & 104.41 & 101.70 \\
    Period 16 & 158.59 & 154.36 & 156.33 & 153.33 & 144.80 & 106.31 & 112.48 & 107.77 \\
    Period 17 & 170.12 & 165.58 & 167.70 & 165.07 & 156.05 & 108.03 & 113.68 & 109.03 \\
    Period 18 & 177.19 & 171.24 & 174.67 & 171.93 & 162.04 & 109.45 & 116.17 & 110.38 \\
    Period 19 & 179.84 & 177.72 & 177.28 & 174.50 & 153.20 & 104.49 & 110.33 & 105.37 \\
    Period 20 & 188.32 & 176.70 & 180.31 & 177.48 & 153.52 & 109.10 & 114.29 & 109.97 \\
    \bottomrule
    \end{tabular}%
  \label{tab:PeriodNTES}%
\end{table}%
\begin{table}[htbp]
  \tiny
  \setlength{\abovecaptionskip}{0pt}
  \setlength{\belowcaptionskip}{5pt}
  \centering
  \caption{The detailed movement of invested \$100 for last 10 days period (Period 20, Stock: BIDU)}
    \begin{tabular}{c|cc|cc|cc|cc|rrrr}
    \toprule
          & \multicolumn{8}{c|}{\textbf{Buy\&Sell decisions by prediction models}} & \multicolumn{4}{c}{\textbf{Buy\&Hold stock/index}} \\
    \midrule
    \multicolumn{1}{c|}{\multirow{2}[0]{*}{\textbf{Day} }} & \multicolumn{2}{c|}{\textbf{MV-GPR}} & \multicolumn{2}{c|}{{\textbf{MV-TPR}}} & \multicolumn{2}{c|}{\textbf{GPR}} & \multicolumn{2}{c|}{\textbf{TPR}} & \multicolumn{1}{c}{\multirow{2}[0]{*}{\textbf{BIDU}}} & \multicolumn{1}{c}{\multirow{2}[0]{*}{\textbf{INDU}}} & \multicolumn{1}{c}{\multirow{2}[0]{*}{\textbf{NDX}}} & \multicolumn{1}{c}{\multirow{2}[0]{*}{\textbf{SPX}}} \\
    \cmidrule(r){2-9}
    \multicolumn{1}{c|}{} & \textbf{Act} & \textbf{Dollar} & \textbf{Act} & \textbf{Dollar} & \textbf{Act} & \textbf{Dollar} & \textbf{Act} & \textbf{Dollar} & \multicolumn{1}{c}{} & \multicolumn{1}{c}{} & \multicolumn{1}{c}{} & \multicolumn{1}{c}{} \\
    \midrule
    \multicolumn{1}{c|}{190} & \multicolumn{2}{c|}{260.24} & \multicolumn{2}{c|}{223.38} & \multicolumn{2}{c|}{264.59} & \multicolumn{2}{c|}{230.23} & 134.23 & 104.49 & 110.33 & 105.37 \\
    \midrule
    \multicolumn{1}{c|}{191} & Buy   & 263.29 & Buy   & 226.00 & Buy   & 267.69 & Buy   & 232.93 & 136.60 & 106.25 & 112.37 & 107.51 \\
    \multicolumn{1}{c|}{192} & Keep  & 272.74 & Keep  & 234.11 & Keep  & 277.29 & Keep  & 241.29 & 141.50 & 108.83 & 115.14 & 110.09 \\
    \multicolumn{1}{c|}{193} & Keep  & 275.50 & Keep  & 236.48 & Keep  & 280.10 & Keep  & 243.73 & 142.94 & 108.99 & 115.52 & 110.60 \\
    \multicolumn{1}{c|}{194} & Keep  & 275.94 & Keep  & 236.85 & Keep  & 280.55 & Keep  & 244.12 & 143.16 & 109.94 & 115.84 & 111.02 \\
    \multicolumn{1}{c|}{195} & Sell  & 275.70 & Sell  & 236.65 & Sell  & 280.31 & Sell  & 243.91 & 142.28 & 110.33 & 115.45 & 111.21 \\
    \multicolumn{1}{c|}{196} & Buy   & 273.26 & Buy   & 234.55 & Buy   & 277.83 & Buy   & 241.75 & 140.95 & 110.37 & 115.55 & 111.20 \\
    \multicolumn{1}{c|}{197} & Keep  & 277.87 & Keep  & 238.52 & Keep  & 282.52 & Keep  & 245.83 & 143.33 & 110.51 & 116.39 & 111.56 \\
    \multicolumn{1}{c|}{198} & Keep  & 272.35 & Keep  & 233.77 & Keep  & 276.90 & Keep  & 240.94 & 140.48 & 110.42 & 116.35 & 111.66 \\
    \multicolumn{1}{c|}{199} & Sell  & 270.29 & Keep  & 233.57 & Sell  & 274.81 & Sell  & 239.13 & 140.36 & 110.08 & 115.53 & 111.11 \\
    \multicolumn{1}{c|}{200} & Keep  & 270.29 & Sell  & 233.29 & Keep  & 274.81 & Keep  & 239.13 & 139.12 & 109.10 & 114.29 & 109.97 \\
    \bottomrule
    \end{tabular}%
  \label{tab:StockBIDU}%
\end{table}%
\begin{table}[htbp]
  \tiny
  \setlength{\abovecaptionskip}{0pt}
  \setlength{\belowcaptionskip}{5pt}
  \centering
  \caption{The detailed movement of invested \$100 for last 10 days period (Period 20, Stock: CTRP)}
    \begin{tabular}{c|cc|cc|cc|cc|rrrr}
    \toprule
          & \multicolumn{8}{c|}{\textbf{Buy\&Sell decisions by prediction models}} & \multicolumn{4}{c}{\textbf{Buy\&Hold stock/index}} \\
    \midrule
    \multicolumn{1}{c|}{\multirow{2}[0]{*}{\textbf{Day} }} & \multicolumn{2}{c|}{\textbf{MV-GPR}} & \multicolumn{2}{c|}{{\textbf{MV-TPR}}} & \multicolumn{2}{c|}{\textbf{GPR}} & \multicolumn{2}{c|}{\textbf{TPR}} & \multicolumn{1}{c}{\multirow{2}[0]{*}{\textbf{CTRP}}} & \multicolumn{1}{c}{\multirow{2}[0]{*}{\textbf{INDU}}} & \multicolumn{1}{c}{\multirow{2}[0]{*}{\textbf{NDX}}} & \multicolumn{1}{c}{\multirow{2}[0]{*}{\textbf{SPX}}} \\
    \cmidrule(r){2-9}
    \multicolumn{1}{c|}{} & \textbf{Act} & \textbf{Dollar} & \textbf{Act} & \textbf{Dollar} & \textbf{Act} & \textbf{Dollar} & \textbf{Act} & \textbf{Dollar} & \multicolumn{1}{c}{} & \multicolumn{1}{c}{} & \multicolumn{1}{c}{} & \multicolumn{1}{c}{} \\
    \midrule
    \multicolumn{1}{c|}{190} & \multicolumn{2}{c|}{169.89} & \multicolumn{2}{c|}{201.99} & \multicolumn{2}{c|}{163.80} & \multicolumn{2}{c|}{176.73} & 80.69 & 104.49 & 110.33 & 105.37 \\
    \midrule
    \multicolumn{1}{c|}{191} & Buy   & 175.02 & Buy   & 208.09 & Buy   & 168.75 & Buy   & 182.07 & 83.65 & 106.25 & 112.37 & 107.51 \\
    \multicolumn{1}{c|}{192} & Keep  & 180.77 & Keep  & 214.92 & Keep  & 174.28 & Keep  & 188.05 & 86.39 & 108.83 & 115.14 & 110.09 \\
    \multicolumn{1}{c|}{193} & Keep  & 185.40 & Keep  & 220.43 & Keep  & 178.75 & Keep  & 192.87 & 88.61 & 108.99 & 115.52 & 110.60 \\
    \multicolumn{1}{c|}{194} & Sell  & 184.91 & Sell  & 220.28 & Sell  & 178.28 & Sell  & 192.36 & 88.55 & 109.94 & 115.84 & 111.02 \\
    \multicolumn{1}{c|}{195} & Keep  & 184.91 & Buy   & 222.82 & Keep  & 178.28 & Keep  & 192.36 & 88.45 & 110.33 & 115.45 & 111.21 \\
    \multicolumn{1}{c|}{196} & Keep  & 184.91 & Keep  & 222.82 & Keep  & 178.28 & Keep  & 192.36 & 88.22 & 110.37 & 115.55 & 111.20 \\
    \multicolumn{1}{c|}{197} & Buy   & 185.16 & Keep  & 222.82 & Buy   & 178.52 & Buy   & 192.62 & 88.92 & 110.51 & 116.39 & 111.56 \\
    \multicolumn{1}{c|}{198} & Keep  & 184.06 & Keep  & 222.82 & Keep  & 177.46 & Keep  & 191.47 & 88.39 & 110.42 & 116.35 & 111.66 \\
    \multicolumn{1}{c|}{199} & Sell  & 183.25 & Keep  & 222.82 & Sell  & 176.67 & Sell  & 190.63 & 88.45 & 110.08 & 115.53 & 111.11 \\
    \multicolumn{1}{c|}{200} & Keep  & 183.25 & Keep  & 222.82 & Keep  & 176.67 & Keep  & 190.63 & 89.20 & 109.10 & 114.29 & 109.97 \\
    \bottomrule
    \end{tabular}%
  \label{tab:StockCTRP}%
\end{table}%
\begin{table}[htbp]
  \tiny
  \setlength{\abovecaptionskip}{0pt}
  \setlength{\belowcaptionskip}{5pt}
  \centering
  \caption{The detailed movement of invested \$100 for last 10 days period (Period 20, Stock: NTES)}
    \begin{tabular}{c|cc|cc|cc|cc|rrrr}
    \toprule
          & \multicolumn{8}{c|}{\textbf{Buy\&Sell decisions by prediction models}} & \multicolumn{4}{c}{\textbf{Buy\&Hold stock/index}} \\
    \midrule
    \multicolumn{1}{c|}{\multirow{2}[0]{*}{\textbf{Day} }} & \multicolumn{2}{c|}{\textbf{MV-GPR}} & \multicolumn{2}{c|}{\textbf{MV-TPR}} & \multicolumn{2}{c|}{\textbf{GPR}} & \multicolumn{2}{c|}{\textbf{TPR}} & \multicolumn{1}{c}{\multirow{2}[0]{*}{\textbf{NTES}}} & \multicolumn{1}{c}{\multirow{2}[0]{*}{\textbf{INDU}}} & \multicolumn{1}{c}{\multirow{2}[0]{*}{NDX}} & \multicolumn{1}{c}{\multirow{2}[0]{*}{SPX}} \\
    \cmidrule(r){2-9}
    \multicolumn{1}{c|}{} & \textbf{Act} & \textbf{Dollar} & \textbf{Act} & \textbf{Dollar} & \textbf{Act} & \textbf{Dollar} & \textbf{Act} & \textbf{Dollar} & \multicolumn{1}{c}{} & \multicolumn{1}{c}{} & \multicolumn{1}{c}{} & \multicolumn{1}{c}{} \\
    \midrule
    \multicolumn{1}{c|}{190} & \multicolumn{2}{c|}{179.84} & \multicolumn{2}{c|}{177.72} & \multicolumn{2}{c|}{177.28} & \multicolumn{2}{c|}{174.50} & 153.20 & 104.49 & 110.33 & 105.37 \\
    \midrule
    \multicolumn{1}{c|}{191} & Buy   & 176.57 & Buy   & 174.49 & Buy   & 174.06 & Buy   & 171.33 & 151.35 & 106.25 & 112.37 & 107.51 \\
    \multicolumn{1}{c|}{192} & Keep  & 184.84 & Keep  & 182.66 & Keep  & 182.21 & Keep  & 179.36 & 158.44 & 108.83 & 115.14 & 110.09 \\
    \multicolumn{1}{c|}{193} & Keep  & 184.16 & Keep  & 181.98 & Keep  & 181.54 & Keep  & 178.69 & 157.86 & 108.99 & 115.52 & 110.60 \\
    \multicolumn{1}{c|}{194} & Keep  & 185.46 & Keep  & 183.27 & Keep  & 182.82 & Keep  & 179.95 & 158.97 & 109.94 & 115.84 & 111.02 \\
    \multicolumn{1}{c|}{195} & Sell  & 185.33 & Keep  & 179.25 & Sell  & 182.69 & Sell  & 179.83 & 155.49 & 110.33 & 115.45 & 111.21 \\
    \multicolumn{1}{c|}{196} & Buy   & 187.22 & keep  & 180.86 & Buy   & 184.55 & Buy   & 181.66 & 156.88 & 110.37 & 115.55 & 111.20 \\
    \multicolumn{1}{c|}{197} & Keep  & 187.75 & Keep  & 181.38 & Keep  & 185.08 & Keep  & 182.18 & 157.33 & 110.51 & 116.39 & 111.56 \\
    \multicolumn{1}{c|}{198} & Sell  & 188.32 & Keep  & 177.52 & Keep  & 181.15 & Keep  & 178.30 & 153.98 & 110.42 & 116.35 & 111.66 \\
    \multicolumn{1}{c|}{199} & Keep  & 188.32 & Sell  & 176.70 & Sell  & 180.31 & Sell  & 177.48 & 153.29 & 110.08 & 115.53 & 111.11 \\
    \multicolumn{1}{c|}{200} & Keep  & 188.32 & Keep  & 176.70 & Keep  & 180.31 & Keep  & 177.48 & 153.52 & 109.10 & 114.29 & 109.97 \\
    \bottomrule
    \end{tabular}%
  \label{tab:StockNTES}%
\end{table}%
\section{Final Investment Details of Stocks in Dow 30}\label{Section:FinalInvestment}
\setcounter{table}{0}
\renewcommand{\thetable}{C\arabic{table}}

\begin{table}[htbp]
  \tiny
  \setlength{\abovecaptionskip}{0pt}
  \setlength{\belowcaptionskip}{5pt}
  \centering
  \caption{The detailed stock investment results under different strategies}
  \begin{tabular}{c|c|c|c|c|c|c|c|c|c}
    \toprule
    \multicolumn{1}{c|}{\multirow{2}[0]{*}{\textbf{Ticker}}} & \multicolumn{1}{c|}{\multirow{2}[0]{*}{\textbf{Industry}}} & \multicolumn{4}{c|}{\textbf{Buy\&Sell Strategy}} & \multicolumn{4}{c}{\textbf{Buy\&Hold Stragegy}} \\
    \cmidrule(r){3-10}
    \multicolumn{1}{c|}{} & \multicolumn{1}{c|}{} & \textbf{MV-GPR} & \textbf{MV-TPR} & \textbf{GPR}   & \textbf{TPR}   & \textbf{Stock} & \textbf{INDU}  & \textbf{NDX}   & \textbf{SPX} \\
    \midrule
    CVX   & Oil \& Gas & 134.97 & 133.69 & 143.47 & 143.81 & 99.38 & \multicolumn{1}{c|}{\multirow{28}[0]{*}{109.10}} & \multicolumn{1}{c|}{\multirow{28}[0]{*}{114.29}} & \multicolumn{1}{c|}{\multirow{28}[0]{*}{109.97}} \\
    XOM   & Oil \& Gas & 128.39 & 132.72 & 131.31 & 136.02 & 99.74 & \multicolumn{1}{c|}{} & \multicolumn{1}{c|}{} & \multicolumn{1}{c|}{} \\
        \cmidrule(r){1-7}
    MMM   & Industrials & 166.76 & 162.96 & 167.12 & 162.65 & 125.96 & \multicolumn{1}{c|}{} & \multicolumn{1}{c|}{} & \multicolumn{1}{c|}{} \\
    BA    & Industrials & 160.12 & 159.98 & 158.60 & 157.38 & 106.39 & \multicolumn{1}{c|}{} & \multicolumn{1}{c|}{} & \multicolumn{1}{c|}{} \\
    CAT   & Industrials & 142.58 & 138.45 & 146.13 & 151.75 & 97.16 & \multicolumn{1}{c|}{} & \multicolumn{1}{c|}{} & \multicolumn{1}{c|}{} \\
    GE    & Industrials & 137.51 & 134.63 & 135.35 & 139.72 & 101.15 & \multicolumn{1}{c|}{} & \multicolumn{1}{c|}{} & \multicolumn{1}{c|}{} \\
    UTX   & Industrials & 144.29 & 139.47 & 143.29 & 145.18 & 101.94 & \multicolumn{1}{c|}{} & \multicolumn{1}{c|}{} & \multicolumn{1}{c|}{} \\
        \cmidrule(r){1-7}
    KO    & Consumer Goods & 128.11 & 128.59 & 124.88 & 124.52 & 112.47 & \multicolumn{1}{c|}{} & \multicolumn{1}{c|}{} & \multicolumn{1}{c|}{} \\
    MCD   & Consumer Goods & 120.69 & 117.09 & 122.19 & 119.59 & 98.81 & \multicolumn{1}{c|}{} & \multicolumn{1}{c|}{} & \multicolumn{1}{c|}{} \\
    PG    & Consumer Goods & 126.62 & 123.32 & 127.14 & 127.10 & 117.04 & \multicolumn{1}{c|}{} & \multicolumn{1}{c|}{} & \multicolumn{1}{c|}{} \\
        \cmidrule(r){1-7}
    JNJ   & Health Care & 146.00 & 146.70 & 147.42 & 145.16 & 113.65 & \multicolumn{1}{c|}{} & \multicolumn{1}{c|}{} & \multicolumn{1}{c|}{} \\
    MRK   & Health Care & 129.40 & 134.48 & 129.36 & 135.05 & 102.45 & \multicolumn{1}{c|}{} & \multicolumn{1}{c|}{} & \multicolumn{1}{c|}{} \\
    PFE   & Health Care & 128.60 & 136.53 & 130.26 & 134.48 & 100.16 & \multicolumn{1}{c|}{} & \multicolumn{1}{c|}{} & \multicolumn{1}{c|}{} \\
    UNH   & Health Care & 164.98 & 164.63 & 166.14 & 162.79 & 131.14 & \multicolumn{1}{c|}{} & \multicolumn{1}{c|}{} & \multicolumn{1}{c|}{} \\
       \cmidrule(r){1-7}
    HD    & Consumer Services & 171.46 & 165.74 & 169.55 & 170.18 & 133.33 & \multicolumn{1}{c|}{} & \multicolumn{1}{c|}{} & \multicolumn{1}{c|}{} \\
    NKE   & Consumer Services & 147.17 & 146.13 & 142.36 & 148.26 & 122.27 & \multicolumn{1}{c|}{} & \multicolumn{1}{c|}{} & \multicolumn{1}{c|}{} \\
    WMT   & Consumer Services & 136.50 & 132.59 & 133.77 & 135.67 & 117.31 & \multicolumn{1}{c|}{} & \multicolumn{1}{c|}{} & \multicolumn{1}{c|}{} \\
    DIS   & Consumer Services & 168.19 & 168.43 & 168.51 & 168.12 & 115.97 & \multicolumn{1}{c|}{} & \multicolumn{1}{c|}{} & \multicolumn{1}{c|}{} \\
        \cmidrule(r){1-7}
    AXP   & Financials & 160.39 & 158.52 & 160.73 & 160.12 & 102.34 & \multicolumn{1}{c|}{} & \multicolumn{1}{c|}{} & \multicolumn{1}{c|}{} \\
    GS    & Financials & 170.46 & 171.29 & 167.71 & 165.72 & 116.16 & \multicolumn{1}{c|}{} & \multicolumn{1}{c|}{} & \multicolumn{1}{c|}{} \\
    JPM   & Financials & 174.90 & 170.07 & 176.48 & 172.12 & 110.09 & \multicolumn{1}{c|}{} & \multicolumn{1}{c|}{} & \multicolumn{1}{c|}{} \\
    TRV   & Financials & 149.81 & 145.88 & 150.70 & 145.71 & 128.18 & \multicolumn{1}{c|}{} & \multicolumn{1}{c|}{} & \multicolumn{1}{c|}{} \\
    V     & Financials & 161.50 & 153.48 & 157.04 & 158.70 & 116.37 & \multicolumn{1}{c|}{} & \multicolumn{1}{c|}{} & \multicolumn{1}{c|}{} \\
        \cmidrule(r){1-7}
    AAPL  & Technology & 201.82 & 206.64 & 203.45 & 208.07 & 147.34 & \multicolumn{1}{c|}{} & \multicolumn{1}{c|}{} & \multicolumn{1}{c|}{} \\
    CSCO  & Technology & 159.13 & 164.88 & 158.61 & 155.92 & 131.34 & \multicolumn{1}{c|}{} & \multicolumn{1}{c|}{} & \multicolumn{1}{c|}{} \\
    IBM   & Technology & 116.10 & 128.79 & 124.92 & 123.74 & 88.06 & \multicolumn{1}{c|}{} & \multicolumn{1}{c|}{} & \multicolumn{1}{c|}{} \\
    INTC  & Technology & 183.80 & 179.45 & 185.52 & 188.22 & 149.24 & \multicolumn{1}{c|}{} & \multicolumn{1}{c|}{} & \multicolumn{1}{c|}{} \\
    MSFT  & Technology & 173.61 & 166.09 & 176.57 & 172.76 & 120.01 & \multicolumn{1}{c|}{} & \multicolumn{1}{c|}{} & \multicolumn{1}{c|}{} \\
    \bottomrule
    \end{tabular}%
  \label{tab:SectorStockDetail}%
\end{table}%
\begin{table}[htbp]
  \tiny
  \setlength{\abovecaptionskip}{0pt}
  \setlength{\belowcaptionskip}{5pt}
  \centering
  \caption{The detailed industry portfolio investment results under different strategies }
    \begin{tabular}{c|cccc|c|c|c|c|}
    \toprule
    \multicolumn{1}{c|}{\multirow{2}[1]{*}{\textbf{Industry Portfolio}}} & \multicolumn{4}{c|}{\textbf{Buy\&Sell Strategy}} & \multicolumn{4}{c|}{\textbf{Buy\&Hold Stragegy}} \\
    \cmidrule(r){2-9}
    \multicolumn{1}{c|}{} & \textbf{MV-GPR} & \textbf{MV-TPR} & \textbf{GPR}   & \textbf{TPR}   & \textbf{Stock} & \textbf{INDU}  & \textbf{NDX}   & \textbf{SPX} \\
    \midrule
    Oil \& Gas & 131.68 & 133.20 & 137.39 & 139.92 & 99.56 & \multicolumn{1}{c|}{\multirow{7}[1]{*}{109.10}} & \multicolumn{1}{c|}{\multirow{7}[1]{*}{114.29}} & \multicolumn{1}{c|}{\multirow{7}[1]{*}{109.97}} \\
    Industrials & 150.25 & 147.10 & 150.10 & 151.34 & 106.52 & \multicolumn{1}{c|}{} & \multicolumn{1}{c|}{} & \multicolumn{1}{c|}{} \\
    Consumer Goods & 125.14 & 123.00 & 124.73 & 123.73 & 109.44 & \multicolumn{1}{c|}{} & \multicolumn{1}{c|}{} & \multicolumn{1}{c|}{} \\
    Health Care & 142.24 & 145.59 & 143.30 & 144.37 & 111.85 & \multicolumn{1}{c|}{} & \multicolumn{1}{c|}{} & \multicolumn{1}{c|}{} \\
    Consumer Services & 155.83 & 153.22 & 153.55 & 155.56 & 122.22 & \multicolumn{1}{c|}{} & \multicolumn{1}{c|}{} & \multicolumn{1}{c|}{} \\
    Financials & 163.41 & 159.85 & 162.53 & 160.47 & 114.63 & \multicolumn{1}{c|}{} & \multicolumn{1}{c|}{} & \multicolumn{1}{c|}{} \\
    Technology & 166.89 & 169.17 & 169.81 & 169.74 & 127.20 & \multicolumn{1}{c|}{} & \multicolumn{1}{c|}{} & \multicolumn{1}{c|}{} \\
    \bottomrule
    \end{tabular}%
  \label{tab:SectorPortfolioDetail}%
\end{table}%

\end{document}